\documentclass[manuscript,nonacm]{acmart}

\AtBeginDocument{%
  \providecommand\BibTeX{{%
    \normalfont B\kern-0.5em{\scshape i\kern-0.25em b}\kern-0.8em\TeX}}}

\usepackage{enumitem}
\begin{document}

\title{Diffusion-Based Visual Art Creation: A Survey and New Perspectives}
% "Collaborative Dynamics Between AI and Human Cognition in Content Creation: An Integrative Approach"
% "Decoding Minds and Emotions: The Role of AI-Generated Content in Understanding Human Cognition"
% "AI-Generated Content: A Dual Perspective on Technological Advancements and Human Cognitive Analysis"
% "From Data to Emotion: Navigating the Interplay of Cognitive Structures and AI-Generated Content"
% "Understanding Human Cognition and Emotion Through AI-Generated Content: A Comprehensive Review"
% Bridging AI Generated Content and Human Cognition: \\

\author{Bingyuan Wang}
\email{bwang667@connect.hkust-gz.edu.cn}
\orcid{0009-0004-3117-3892}
\affiliation{%
  \institution{The Hong Kong University of Science and Technology (Guangzhou)}
  \streetaddress{No. 1, Duxue Road, Dongchong Town, Nansha District.}
  \city{Guangzhou}
  \state{Guangdong}
  \country{China}
  \postcode{511453}
}

\author{Qifeng Chen}
\email{cqf@ust.hk}
\orcid{0000-0003-2199-3948}
\affiliation{%
  \institution{The Hong Kong University of Science and Technology}
  \streetaddress{Clear Water Bay, Kowloon}
  \state{Hong Kong SAR}
  \country{China}
  \postcode{999077}
  }

\author{Zeyu Wang}
\authornote{Corresponding author.}
\email{zeyuwang@ust.hk}
\orcid{0000-0001-5374-6330}
\affiliation{%
  \institution{The Hong Kong University of Science and Technology (Guangzhou)}
  \streetaddress{No. 1, Duxue Road, Dongchong Town, Nansha District.}
  \city{Guangzhou}
  \state{Guangdong}
  \country{China}
  \postcode{511453}
  }

\renewcommand{\shortauthors}{Wang et al.}
% Wang et al.

\begin{abstract}

The integration of generative AI in visual art has revolutionized not only how visual content is created but also how AI interacts with and reflects the underlying domain knowledge. This survey explores the emerging realm of diffusion-based visual art creation, examining its development from both artistic and technical perspectives. We structure the survey into three phases, data feature and framework identification, detailed analyses using a structured coding process, and open-ended prospective outlooks. Our findings reveal how artistic requirements are transformed into technical challenges and highlight the design and application of diffusion-based methods within visual art creation. We also provide insights into future directions from technical and synergistic perspectives, suggesting that the confluence of generative AI and art has shifted the creative paradigm and opened up new possibilities. By summarizing the development and trends of this emerging interdisciplinary area, we aim to shed light on the mechanisms through which AI systems emulate and possibly, enhance human capacities in artistic perception and creativity.

\end{abstract}






\maketitle

\section{Introduction}
\label{sec:introduction}

% Why为什么，motivation——
% 	AIGC在这两年取得了很大的进展和影响，也取得了很好的效果；然而这些方法大多数应用于自然图像（好收集、量大），相对较少的研究关注diffusion-based methods for artistic painting.
% 	art painting很多种，艺术价值高，更creative，大家感兴趣
As an emerging concept and evolving field, Artificial Intelligence Generated Content (AIGC) has made significant progress and impact over the past several years, especially since the diffusion model was proposed \cite{ho2020denoising}. On the other hand, visual art, encompassing a wide variety of genres, media, and styles, possesses high artistic value and diverse creativity, sparking widespread interest. However, compared to general method innovations \cite{song2020denoising, ho2022classifier} and specific model designs \cite{esser2021taming, ramesh2021zero}, relatively limited research focuses on diffusion-based methods for visual art creation. Fewer works thoroughly examine the problem, summarize frameworks, or provide trends and insights for future research.

Relevant surveys approach this problem from both technical and artistic perspectives. Some recent surveys focus on the intersection of artificial intelligence with content generation, examining data modalities and tasks~\cite{li2023generative, foo2023ai} to methodological progressions and applications~\cite{cao2023comprehensive, bie2023renaissance}. These surveys reviewed a series of work on artistic stylization~\cite{kyprianidis2012state}, appearance editing~\cite{schmidt2016state}, text-to-image transitions~\cite{zhang2023text}, and the newfound applications of AI across multiple data modalities~\cite{zhang2023survey}. Methodologically, they span neural style transfer~\cite{jing2019neural}, GAN inversion~\cite{xia2022gan} to attention mechanisms~\cite{guo2022attention} and diffusion models~\cite{chen2024overview}—each contributing to the state of the art in their own right. From an application perspective, they explore the transformative integration of AIGC across various domains, and while remarkable, they also highlight challenges that call for further development and ethical consideration \cite{liu2023transformation,xu2024unleashing}. Meanwhile, surveys with an artistic focus unravel the interplay between arts and humanities within the AIGC era, probing into the processing and understanding of art through advanced computational methods~\cite{bengamra2024comprehensive, castellano2021deep, liu2023neural}, the generative potential of AI in creating novel art forms~\cite{depolo2021after, zhang2021comprehensive}, and the applicability of its integration in enhancing educational and therapeutic experiences \cite{mai2023brain, chen2024empowering}. We noticed a lack of surveys that specifically focus on combining diffusion-based models with visual art creation, and aim to fill this gap with our work.

This survey aims to provide a comprehensive review of the intersection of diffusion-based generative methods and visual art creation. We define the research scope through two independent taxonomies from technical and artistic perspectives, identifying diffusion-based generative techniques as one of the key methods and art as a significant application scenario. Our research goals are to \emph{analyze how diffusion models have revolutionized visual art creation} and to \emph{offer frameworks and insights for future research in this area}. We address four main research questions that explore the trending topics, current challenges, employed methods, and future directions in diffusion-based visual art creation.

We first conduct a structural analysis of diffusion-based visual art creation, which highlights current hot topics and evolving trends. Through categorizing data into application, understanding, and generation, we find a concentration of research on generation, specifically in controllable, application-oriented, and historically- or genre-specific art creation~\cite{choi2021ilvr, wang2023cclap, liu2023portrait}. Furthermore, we present a new analytical framework that aligns artistic scenarios with data modalities and generative tasks, allowing for a structured approach to the research questions.
Temporal analysis suggests a post-diffusion boom in visual art creation, with a steady rise in diffusion-based methods~\cite{saharia2022photorealistic, rombach2022high, gal2022image}. Generative methods have shifted from traditional rule-based simulations to diffusion-model modifications, along with a progression from image-only inputs to more controllable conditional formats, and an increase in dataset generality and process complexity~\cite{ruiz2023dreambooth, zhang2023adding, mou2024t2i, bar2023multidiffusion}.
The emerging trends point towards a technical evolution from basic model frameworks to interactive systems and a shift in focus toward user requirements for creativity and interactivity. These findings set the stage for our survey, which aims to bridge the gap between technological advancements and artistic creation, fostering a synergy that can lead to a new wave of innovation in AIGC.

The exploration from artistic requirements to technical problems forms a cornerstone of our investigation in diffusion-based visual art. We scrutinize the symbiotic relationship between application domains, artistic genres, and their correspondence with data modality and generative tasks. Supported by a robust body of work~\cite{Abrahamsen2023InventingAS, zhang2024artbank, qiao2022initial, huang2022draw, liao2022artbench}, we delve into how different visual art forms and domains drive the development of technical solutions.
From complex artistic scenarios to specific genres like traditional Chinese painting~\cite{lyu2024diffusion, fu2021multi, wang2024intelligent, li2021paint4poem, wang2023magicscroll}, our approach deciphers the intersection of AI and art, translating artistic goals into computational tasks. We establish a framework that defines these relationships via data modalities—ranging from brush strokes~\cite{li2024inverse} to 3D models~\cite{dong2024interactive3d}—and generative tasks like quality enhancement~\cite{he2023scalecrafter} and controllable generation~\cite{kawar2023imagic}.
These tasks are then meticulously tied to artistic objectives, with corresponding evaluation metrics such as CLIP Score for controllability~\cite{radford2021learning, patashnik2021styleclip} and FID for fidelity~\cite{heusel2017gans, salimans2016improved}. This multifaceted evaluation system ensures that the generated art not only meets technical standards but also fulfills the nuanced demands of artistic creation, aligning with the evolving trends in the post-diffusion era.

We also investigate the intricate designs and applications of diffusion-based methods to explore how these methods enhance the generative process in visual art. We offer a detailed classification of tasks such as controllable generation, content editing, and stylization, each bolstered by novel diffusion-based approaches that prioritize user input and artistic integrity~\cite{choi2021ilvr, gal2022image}.
Innovations like ILVR~\cite{choi2021ilvr} and ControlNet~\cite{zhang2023adding} exemplify the strides made in achieving precise control over image attributes, while advances in methods like GLIDE~\cite{nichol2021glide} and InstructPix2Pix~\cite{brooks2023instructpix2pix} showcase the growing sophistication in content editing and the ability to adaptively respond to textual prompts. Stylization techniques, such as InST~\cite{zhang2023inversion} and DiffStyler~\cite{huang2024diffstyler}, demonstrate the nuanced application of artistic styles, while quality enhancement tools like eDiff-I~\cite{balaji2022ediff} and PIXART-$\alpha$~\cite{chen2023pixart} push the limits of image resolution and fidelity.
Furthermore, we categorize these methods based on a unified diffusion model structure, highlighting advancements in individual modules such as encoder-decoders, denoisers, and noise predictors~\cite{lu2023painterly, liu2022compositional, chefer2023attend}. These developments manifest in trends that emphasize attention mechanisms, personalization, control, quality, modularity, multi-tasking, and efficiency~\cite{cao2023masactrl, ruiz2023dreambooth, kumari2023multi}. The synthesis of these trends reflects a dynamic evolution in diffusion-based generative models, marking a transformative era in visual art creation.

The frontiers of diffusion-based visual art creation are seen through the lens of technical evolution and human-AI collaboration. Technically, we are witnessing a leap into higher dimensions and more diverse modalities, transcending traditional boundaries to create immersive experiences~\cite{zhang2022arf, zhang2024coarf}.
A synergistic perspective reveals a future where human and AI collaboration is seamless, allowing for interactive systems that augment human creativity and facilitate a deeper reception and alignment with content~\cite{chung2023promptpaint, dall2023collaborative, guo2023artverse}. 
These approaches range from the use of human concepts as task inspiration to the generation of content that resonates emotionally and models that encapsulate the essence of creativity~\cite{sartori2014affective, yang2024emogen, wu2022creative}. This multidimensional approach is shifting paradigms, enabling a greater understanding and co-creation between humans and AI~\cite{russo2022creative, zheng2024creativeseg, zhong2023let}. It paints a future where the boundaries between human and AI creativity become blurred, leading to a new era of digital artistry.

In summary, our literature review yields the following contributions:
\begin{itemize}
    \item A comprehensive dataset and taxonomy of AIGC techniques in visual art creation, coded with multi-dimensional, fine-grained labels.
    \item A framework for analyzing and categorizing the relationship between diffusion-based generative methods and their applications in visual art creation, with multi-faceted features and relationships as key findings. 
    \item A summary of frontiers, trends, and future outlooks from multiple interdisciplinary perspectives.
\end{itemize}
\section{Background}

Prior to the emergence of diffusion models, the field of machine learning in visual art creation had already gone through several significant developments. These stages were marked by various generative models that opened new chapters in image synthesis and editing.
One of the earliest pivotal advancements was the introduction of Generative Adversarial Networks (GANs) by Goodfellow et al. \cite{goodfellow2014generative}, which introduced a new framework where a generator network learned to produce data distributions through an adversarial process. Following closely, CycleGAN by Zhu et al. \cite{zhu2017unpaired} overcame the need for paired samples, enabling image-to-image translation without paired training samples. These models gained widespread attention due to their potential in a variety of visual content creation tasks.
Simultaneously with the development of GANs, another important class of models was the Variational Autoencoder (VAEs) introduced by Kingma and Welling \cite{kingma2013auto}, which offered a method to generate continuous and diverse samples by introducing a latent space distribution. This laid the groundwork for controllable image synthesis and inspired a series of subsequent works.
With enhanced computational power and innovation in model design, Karras et al. pushed the quality of image generation further with StyleGAN \cite{karras2020analyzing}, a model capable of producing high-resolution and lifelike images, driving more personalized and detailed image generation.
The incorporation of attention mechanisms into generative models significantly improved the relevance and detail of generated content. The Transformer by Vaswani et al. \cite{vaswani2017attention}, with its powerful sequence modeling capabilities, influenced the entire field of machine learning, and in visual art generation, the successful application of Transformer architecture to image recognition with Vision Transformer (ViT) by Dosovitskiy et al. \cite{dosovitskiy2020image}, and further for high-resolution image synthesis with Taming Transformers by Esser et al. \cite{esser2021taming}, showed the immense potential of Transformers in visual generative tasks.
Subsequent developments like SPADE by Park et al. \cite{park2019semantic} and the time-lapse synthesis work by Nam et al. \cite{nam2019end} marked significant steps towards more complex image synthesis tasks. These methods provided richer context awareness and temporal dimension control, offering users more powerful creative expression capabilities.
The introduction of Denoising Diffusion Probabilistic Models (DDPMs) by Ho et al. \cite{ho2020denoising} and the subsequent showcase by Ramesh et al. of DALL·E \cite{ramesh2021zero}, which could create images from textual prompts based on such models, marked another leap forward for generative models, adding a new chapter to the history of model development. These models achieved breakthroughs in image quality and also demonstrated new possibilities in terms of controllability and diversity.
These developments constitute a rich history of visual art creation in the field of machine learning, laying a solid foundation for the arrival of the diffusion era. In this survey, we will delve deeper into how diffusion models inherit and transcend the boundaries of these prior technologies, opening a new chapter in creative generation.

From an artistic perspective, the advancements in machine learning and generative models have intersected intriguingly with the domain of visual arts, which encompasses a wide variety of genres, media, and styles. Artists have traditionally held the reins of creative power, with the ability to produce works that carry significant artistic value and cultural resonance. The introduction of sophisticated generative algorithms offers a new toolkit for artists, potentially expanding the boundaries of their creativity~\cite{mazzone2019art}.
As these technological tools become more accessible and integrated into artistic workflows, they present an opportunity for artists to experiment with novel forms of expression, blending traditional techniques with computational processes~\cite{elgammal2017can}. This fusion sparks widespread interest not only within the tech community but also among art enthusiasts who are curious about the new creative possibilities~\cite{miller2019artist}.
Machine learning models, especially those capable of generating high-quality visual content, are increasingly seen as collaborators in the artistic process. Rather than replacing human creativity, they are enhancing it, enabling artists to explore complex patterns, intricate details, and conceptual depths that were previously difficult or impossible to achieve manually~\cite{gatys2016image}.
This symbiotic relationship between artist and algorithm is transforming the landscape of visual art. Artists are beginning to harness these models to create works that challenge our understanding of art and authorship~\cite{mccormack2019autonomy}. As a result, the dialogue between technology and art is becoming richer, with machine learning models contributing to the creation of art that offers greater creative freedom and artistic value. This evolving dynamic prompts both excitement and philosophical reflection on the nature of creativity and the role of artificial intelligence in the future of artistic expression.

\section{Related Work}
In this section, we provide an overview of the scope of AIGC and contributions of pertinent surveys that concentrate on fields and topics relevant to diffusion-based visual art creation. We first collected 42 surveys and filtered out 30 by relevance. These surveys are primarily categorized by their focus on either technical (17) or artistic (13) aspects. Collectively, they establish the paradigm of this interdisciplinary field and create a platform for our discussion.

\subsection{Relevant Surveys with Technical Focus}

From a technical view, a tier of surveys focus on the advancements and implications of artificial intelligence in content generation. For example, Cao et al.~\cite{cao2023comprehensive} provides a detailed review of the history and recent advances in AIGC, highlighting how large-scale models have improved the extraction of intent information and the generation of digital content such as images, music, and natural language. We further break down this view into data and task, method, and application perspectives.

\subsubsection{Data and Task Perspectives.}
A series of surveys inspect AIGC from data and modality and highlight the evolution and challenges in various tasks, including artistic stylization, appearance editing, text-to-image generation, text-to-3D transformation, and AI-generated content across multiple modalities.
Prior to the diffusion era, the survey by Kyprianidis et al.~\cite{kyprianidis2012state} delves into the field of nonphotorealistic rendering (NPR), presenting a comprehensive taxonomy of artistic stylization techniques for images and video. It traces the development of NPR from early semiautomatic systems to the automated painterly rendering methods driven by image gradient analysis, ultimately discussing the fusion of higher-level computer vision with NPR for artistic abstraction and the evolution of real-time stylization techniques.
Schmidt et al.~\cite{schmidt2016state} review the state of the art in the artistic editing of appearance, lighting, and material, essential for conveying information and mood in various industries. The survey categorizes editing approaches, interaction paradigms, and rendering techniques while identifying open problems to inspire future research in this complex and active area.
In the era of large generative models, Bie et al.'s survey on text-to-image generation (TTI)~\cite{bie2023renaissance} explores how the integration with large language models and the use of diffusion models have revolutionized TTI, bringing it to the forefront of machine-learning research and greatly enhancing the fidelity of generated images. The review provides a critical comparison of existing methods and proposes potential improvements and future pathways, including video and 3D generation.
Li et al. conducted the first comprehensive survey on text-to-3D~\cite{li2023generative}, an active research field due to advancements in text-to-image and 3D modeling technologies. The work introduces 3D data representations and foundational technologies, summarizing how recent developments realize satisfactory text-to-3D results and are used in various applications like avatar and scene generation.
Finally, Foo et al.'s survey on AI-generated content~\cite{foo2023ai} spans a plethora of data modalities, from images and videos to 3D shapes and audio. It reviews single-modality and cross-modality AIGC methods, discusses the representative challenges and works in each modality, and suggests future research directions.

\subsubsection{Method Perspective.}
A main body of recent surveys in generative AI and computer vision has been on the evolution of methodologies for style transfer, GAN inversion, attention mechanisms, and diffusion models, which have been instrumental in driving forward the state-of-the-art.
Neural Style Transfer (NST) has evolved into a field of its own, with a variety of algorithms aimed at improving or extending the seminal work of Gatys et al. ~\cite{jing2019neural} provides a taxonomy of NST algorithms and compares them both qualitatively and quantitatively, also highlighting the potential applications and future challenges in the field.
In the realm of GANs, the survey on GAN inversion~\cite{xia2022gan} details the process of inverting images back into the latent space to enable real image editing and interpreting the latent space of GANs. It outlines representative algorithms, applications, and emerging trends and challenges in this area.
The survey on attention mechanisms in computer vision~\cite{guo2022attention} categorizes them based on their approach, including channel, spatial, temporal, and branch attention. This comprehensive review links the success of attention mechanisms in various visual tasks to the human ability to focus on salient regions in complex scenes, and it suggests future research directions.
Diffusion-based image generation models have seen significant progress, paralleling advancements in large language models like ChatGPT. ~\cite{zhang2023survey} examines the issues and solutions associated with these models, particularly focusing on the stable diffusion framework and its implications for future image generation modeling.
Text-to-image diffusion models are also reviewed~\cite{zhang2023text}, offering a self-contained discussion on how basic diffusion models work for image synthesis. This includes a review of state-of-the-art methods on text-conditioned image synthesis, applications beyond, and existing challenges.
Retrieval-Augmented Generation (RAG) for AIGC is discussed in a survey that classifies RAG foundations and suggests future directions by illuminating advancements and pivotal technologies~\cite{zhao2024retrieval}. The survey provides a unified perspective encompassing all RAG scenarios, summarizing enhancement methods, and surveying practical applications across different modalities and tasks.
Finally, an overview of diffusion models addresses their applications, guided generation, statistical rates, and optimization~\cite{chen2024overview}. It reviews emerging applications and theoretical aspects of diffusion models, exploring their statistical properties, sampling capabilities, and new avenues in high-dimensional structured optimization.

\subsubsection{Application Perspective.}
From an application perspective, recent surveys have explored the integration and impact of AIGC across different domains such as brain-computer interfaces, education, and mobile networks, emphasizing its transformative potential.
Mai et al.'s survey~\cite{mai2023brain} introduces the concept of Brain-conditional Multimodal Synthesis within the AIGC framework, termed AIGC-Brain. This domain leverages brain signals as a guiding condition for content synthesis across various modalities, aiming to decode these signals back into perceptual experiences. The survey provides a detailed taxonomy for AIGC-Brain decoding models, task-specific implementations, and quality assessments, offering insights and prospects for research in brain-computer interface systems.
Chen's systematic literature review~\cite{chen2024empowering} addresses AIGC's application in education, highlighting the profound impact of technologies like ChatGPT. The review identifies key themes such as performance assessment, instructional applications, and the advantages and risks of AIGC in education. It delves into the research trends, geographical distribution, and future agendas to integrate AI more effectively into educational methods, tools, and innovation.
Xu et al.~\cite{xu2024unleashing} survey the deployment of AIGC services in mobile networks, focusing on providing personalized and customized content while preserving user privacy. The survey examines the lifecycle of AIGC services, collaborative cloud-edge-mobile infrastructure, creative applications, and the associated challenges of implementation, security, and privacy. It also outlines future research directions for enhancing mobile AIGC networks.

These technically oriented surveys characterize remarkable advancements in the field of generative AI, emphasizing the innovative algorithms and interaction paradigms that enable the creation of diverse content across various data modalities. However, they also point out the existing challenges, including the need for further technical development, the consideration of ethical issues, and the imperative to address potential negative impacts on society. 

\subsection{Relevant Surveys with Artistic Focus}

Another tier of work adopts an artistic view by specifically focusing on arts and humanities in the AIGC era. For example, Liu et al.~\cite{liu2023transformation} explore the transformational impact of artificial general intelligence (AGI) on the arts and humanities, addressing critical concerns related to factuality, toxicity, biases, and public safety, and proposing strategies for responsible deployment. We further break the view into processing and understanding, generation, and application perspectives.

\subsubsection{Processing and Understanding Perspectives.}
The surveys with an artistic focus shed light on the intersection of art and technology, where advanced processing techniques and computational methods are employed to understand and enhance the appreciation of visual arts.
Depolo et al.'s review~\cite{depolo2021after} discusses the mechanical properties of artists' paints, emphasizing the importance of understanding paint material responses to stress through tensile testing data and other innovative techniques. The study highlights how new methods allow for the investigation of historic samples with minimal intervention, utilizing techniques such as nanoindentation, optical methods like laser shearography, computational simulations, and non-invasive approaches to predict paint behavior.
Castellano et al.~\cite{castellano2021deep} provides an overview of deep learning in pattern extraction and recognition within paintings and drawings, showcasing how these technological advances paired with large digitized art collections can assist the art community. The goal is to foster a deeper understanding and accessibility of visual arts, promoting cultural diffusion.
Zhang et al.'s comprehensive survey on the computational aesthetic evaluation of visual art images~\cite{zhang2021comprehensive} tackles the challenge of quantifying aesthetic perception. It reviews various approaches, from handcrafted features to deep learning techniques, and explores applications in image enhancement and automatic generation of aesthetic-guided art, while addressing the challenges and future directions in this field.
Liu et al.'s review on neural networks for hyperspectral imaging (HSI) of historical paintings~\cite{liu2023neural} details the application of neural networks for pigment identification and classification. By focusing on processing large spectral datasets, the review contributes to the application of these networks, enhancing artwork analysis and preservation of cultural heritage.
Lastly, Bengamra's survey on object detection in visual art~\cite{bengamra2024comprehensive} offers a taxonomy of the methods used in the analysis of artwork images, proposing a classification based on the degree of learning supervision, methodology, and style. It outlines challenges and future directions for improving object detection performance in visual art, contributing to the overall understanding of human history and culture through art.

\subsubsection{Generation Perspective.}
Surveys focused on the generation of art through AI technologies underscore the transformative role AI plays in both understanding and creating visual arts. 
Cetinic et al.~\cite{cetinic2022understanding} offer an integrated review of AI's dual application in art analysis and creation, including an overview of artwork datasets and recent works tackling various tasks such as classification and computational aesthetics, as well as practical and theoretical considerations in the generation of AI Art.
Shahriar et al.~\cite{shahriar2022gan} examine the potential of GANs in art creation, exploring their use in generating visual arts, music, and literary texts. This survey highlights the performance and architecture of GANs, alongside the challenges and future recommendations in the field of computer-generated arts.
Liu's overview of AI in painting~\cite{liu2023overview} reveals the field's current status and future direction, discussing how AI algorithms can produce unique art forms and automate tasks in traditional painting, thereby promising a revolution in the digital art world and traditional painting processes.
Ko et al.~\cite{ko2023large} delve into Large-scale Text-to-Image Generation Models (LTGMs) like DALL-E, discussing their potential to support visual artists in creative works through automation, exploration, and mediation. The study includes an interview and literature review, offering design guidelines for future intelligent user interfaces using LTGMs.
Lastly, Maerten et al.'s review on deep neural networks in AI-generated art~\cite{maerten2023paintbrush} examines the evolution of these architectures, from classic convolutional networks to advanced diffusion models, providing a comparison of their capabilities in producing AI-generated art. This review encapsulates the rapid progress and interaction between art and computer science.

\subsubsection{Application Perspective.}
The surveys with an artistic focus on application delve into the transformative potential of integrating art with other disciplines, particularly science education and therapy, to foster holistic learning and healing experiences.
Turkka et al.~\cite{turkka2017integrating} investigates how art is integrated into science education, revealing through a qualitative e-survey of science teachers (n=66) that while the incorporation of art can enhance teaching, it is infrequently applied in classroom practices. The study presents a pedagogical model for art integration, which characterizes integration through content and activities, and suggests that teacher education should provide more consistent opportunities for art integration to enrich science teaching.
The study on art therapy~\cite{hu2021art} surveys the clinical applications and outcomes of art therapy as a non-pharmacological intervention for mental disorders. The systematic review of 413 literature pieces underscores the clinical effectiveness of art therapy in alleviating symptoms of various mental health conditions, such as depression, anxiety, and cognitive impairments, including Alzheimer’s and autism. It emphasizes the therapeutic power of art in assisting patients to express emotions and providing medical specialists with complementary diagnostic information.

The artistically-oriented surveys reveal how technological advancements—specifically in AI—have revolutionized not only the analysis and preservation of visual arts but also enabled the active creation of innovative art forms. These studies underscore the potential of AI to deeply understand artistic nuances and contribute creatively, thus enriching the artistic domain with new tools and methodologies.
% 对相关综述的分析 relevant surveys
% Motivation: existing surveys in diffusion models and visual art creation, but not enough work to combine both perspectives. 
However, we observe that while there is existing literature surveying diffusion models and visual art creation individually, there is a gap in research that synthesizes both perspectives. This opens up an opportunity to explore how diffusion models can be specifically applied to the domain of visual art creation, potentially leading to innovative approaches that could transform content production and understanding. We aim to bridge this gap by merging the technical intricacies of generative AI with the creative process of art. By doing so, we seek to contribute to the ongoing dialogue between art and technology, enhancing the creative process and expanding the scope of possibilities in visual art creation.

\section{Research Scope and Concepts}
In this section, we first define the survey's research scope and explain relevant concepts. Then, we summarize our research goals and target questions. Together, they establish a coherent context and lay a foundation for the following sections.

\subsection{Research Scope}
\label{sec: scope}
\begin{figure}
    \centering
    \includegraphics[width=0.8\textwidth]{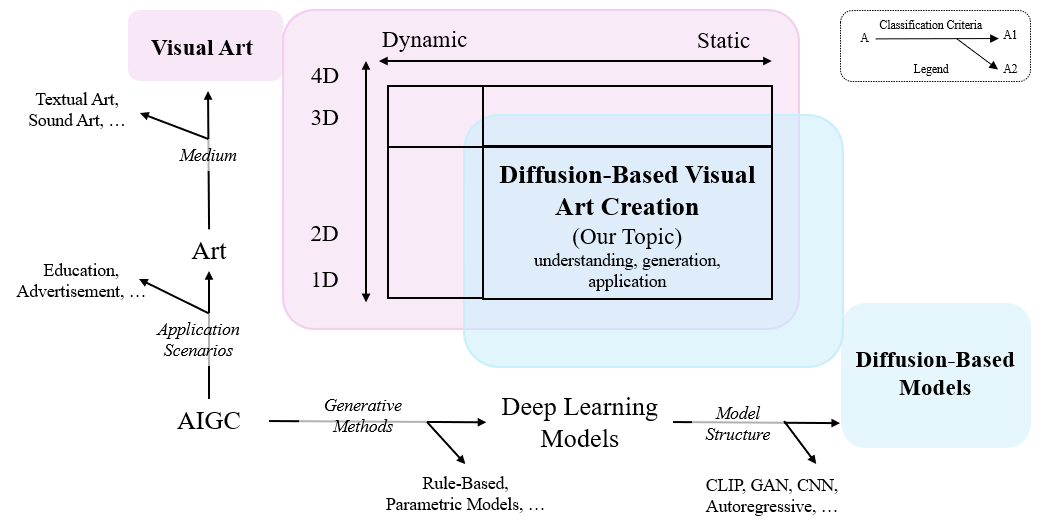}
    \caption{Identifying the scope of this survey. We adopt two independent taxonomies to determine the research scope. For visual arts (creative targets), we primarily include 2D static visual content, supplemented by a small amount of animation, 3D, and cartoons. Regarding diffusion models (generative methods), we mainly cover aspects such as model design, task applications, and human-computer interaction.}
    \label{fig:framework_0}
\end{figure}

Based on the surveys discussed in the previous section, we identify two independent taxonomies in the technical and artistic realms. The first taxonomy, typical in surveys with a technical focus, categorizes diffusion-based generative techniques as one of the generative methods and art as an application scenario~\cite{cao2023comprehensive, li2023generative, foo2023ai, liu2023transformation}.
On the other hand, surveys with an artistic stance commonly adopt historical or theoretical perspectives, categorize relevant research by application scenarios and artistic categories (in Sec.~\ref{sec: framework}, we correspond them to different data modalities or applications), and focus more on the implications of generated results~\cite{liu2023overview, liu2023neural, depolo2021after, castellano2021brief, zhang2021comprehensive}.
We display the two taxonomies and our research scope in Fig.~\ref{fig:framework_0}. The independent taxonomies are represented as perpendicular axes. Following our motivation, this survey lies in the intersection of these two axes.

% 右上角太空

\subsection{Relevant Concepts}
To clearly define our research scope and differentiate it from similar work, we provide an explanation and categorization method for the two most relevant concept realms and their sub-concepts.

\subsubsection{Diffusion Model.} In Jan. 2020, Ho et al. proposed Denoising Diffusion Probabilistic Models~\cite{ho2020denoising} and tested its performance on multiple image synthesis tasks, proclaiming the advent of the post-diffusion era. Ten months later, Song et al. adapted the denoising process to the latent space and significantly improved the generative performance, which is called Denoising Diffusion Implicit Models~\cite{song2020denoising}. In 2021, different researchers optimized the method by integrating advanced text-image encoders (e.g., CLIP~\cite{radford2021learning}) and conditioning methods (e.g., ILVR~\cite{choi2021ilvr}). Another series of work systematically framed the generative task~\cite{nichol2021glide} and established relevant benchmarks~\cite{dhariwal2021diffusion}, demonstrating surpassing performance than previous state-of-the-art methods. 

In early 2022, many technical companies released respective diffusion-based generative frameworks, including DALL·E-2~\cite{ramesh2022hierarchical}, Imagen~\cite{saharia2022photorealistic}, Stable Diffusion~\cite{rombach2022high}, etc. These methods feature extensive training and can generate high-quality, artistic images to meet commercial needs. From late 2022, the field has shifted from a common focus to different sub-tracks and downstream applications, by diversifying multiple tasks, introducing different methods, and adapting to various scenarios. Meanwhile, within the AIGC framework, the field of Natural Language Processing (NLP) has also witnessed significant breakthroughs. Researchers proposed foundational models (e.g., the GPT series~\cite{openai_model_index}), designed adaptation methods (e.g., LoRA~\cite{hu2021lora}), and achieved comparable performance with humans in NLP tasks~\cite{bubeck2023sparks}. Combined with these advancements, the field of Diffusion Model increased in both width and inclusiveness, becoming more expanded and more interconnected with other fields.

\begin{figure}
    \centering
    \includegraphics[width=\textwidth]{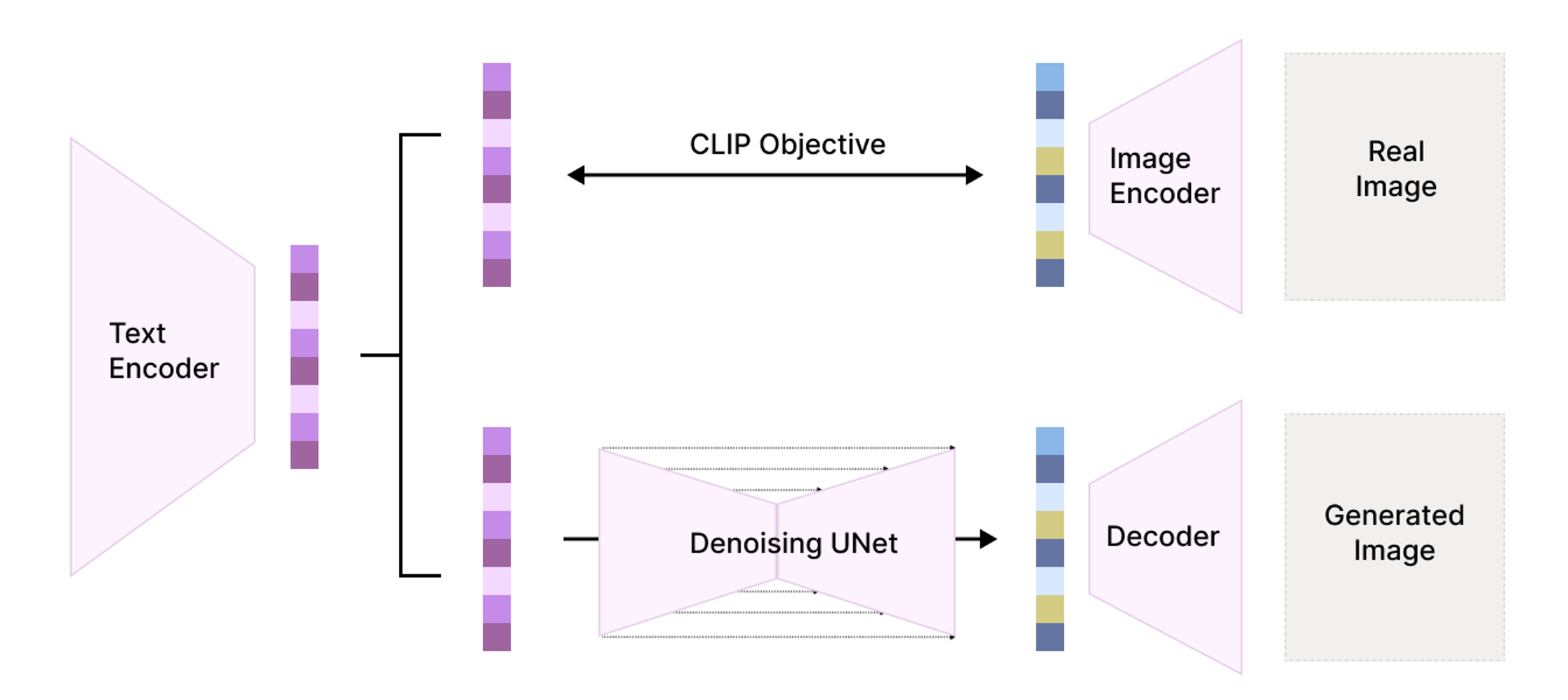}
    \caption{Diffusion-based generative structures suggested by Stable Diffuson~\cite{rombach2022high} and DALL·E-2~\cite{ramesh2022hierarchical}. The image illustrates how the diffusion model integrates with the CLIP model to form the pipeline for generative tasks. The upper half shows the training process, and the lower half shows the inference process with the internal mechanism of diffusion models.
    }
    \label{fig:model_structure}
\end{figure}

Fig.~\ref{fig:model_structure} illustrates the basic structure of how a diffusion model is used as the core structure for a complete generative process (Fig.~\ref{fig:model_structure}-(a)), and how the model is combined with other pre-trained foundational models (such as CLIP as text-image encoder, Fig.~\ref{fig:model_structure}-(b)) to accomplish typical text-to-image generative tasks. From a technical perspective, the structure of a diffusion-based generative model typically consists of the following five parts:  
\begin{itemize}
    \item \textbf{Encoder and Decoder.} In image generation, the encoder and decoder connect the pixel space and latent space. During the generation process, the encoder compresses the input data into a latent representation, and the decoder subsequently reconstructs the output from this compressed form.
    \item \textbf{Denoiser.} As a core component, the denoiser works to remove noise from the latent, in a step-by-step manner, by a set of learned Gaussian noises. Researchers design both new model structures and denoising processes for better performance.
    \item \textbf{Noise Predictor.} This module predicts key parameters of noise distribution, which is learned during the training process. Setting proper noise can guide the generation process toward intended targets.
    \item \textbf{Post-Processor.} After the initial output is generated, this module refines the results by enhancing its resolution and final quality.
    \item \textbf{Additional Modules.} These may include any extra components that supplement the core functionality, such as modules to improve controllability or fulfill specific tasks.
\end{itemize}

In Sec.~\ref{sec: method_design}, we will adhere to this framework to categorize different generative methods. 
\label{sec: model_structure}

\subsubsection{Visual Art.} 

% In Fig.~\ref{fig:visual_art}, we display multiple forms of visual art. 
We break up the topic by three different perspectives of visual art 1) as a conceptual realm under \emph{art}, 2) as visual contents created by \emph{artists}, and 3) as generated results with the quality of \emph{artistic}. Such perspectives will be revisited in the Discussion section (Sec.~\ref{sec: discussion}).
\begin{itemize}
    \item \textbf{What is Visual Art?} Art is broadly understood to encompass a variety of creative expressions that convey ideas, emotions, and concepts through various media. Among them, Visual Art features different visual forms as media~\cite{lazzari2008exploring}. This field can be further split into different artistic categories, including painting, sketch, sculpture, etc. (where painting is the most common category), and different dimensions, including 1D brush stroke, 2D images, 3D scenes, etc. (where 2D images are the most common representation). 
    \item \textbf{Who is the Artist?} Throughout our history, different schools of criticism have adopted different views on the subject and creation of artwork. One perspective of AIGC is to mimic the role of artists with advanced generative models, which provides a possible future framework for creativity~\cite{mccormack2014future}. 
    \item \textbf{How is Visual Art Created?} In visual art generation, a common way is to incorporate human aesthetics and expertise into both the generative and evaluation processes. Such a system includes data choice, task definition, and model design, which is parallel to the artist's role as both creators and receivers (e.g., ~\cite{elgammal2017can}). However, the process is data-driven and rule-based, which is different from perception, emotion, and creativity as artists' driving forces. 
    \item \textbf{What is Artistic?} In the artistic realm, the definition of artistic features more psychological and philosophical elements and is also under debate~\cite{dutton2009art}. However, scientific researchers often adopt a common ground and propose more acceptable standards and more approachable metrics. As will be discussed in Sec.~\ref{sec: analysis}, the commonly referred terms include high quality, stylistic/realistic, controllable, etc.
    
    % The two different dimensions of artistic (a technical standard and an art standard)   
    
\end{itemize}

% \begin{figure}
%     \centering
%     \includegraphics[width=\textwidth]{fig/visual art.png}
%     \caption{Different forms of AI-generated visual art. (a) Generating images in a certain artistic category with surpassing details (e.g., ink painting~\cite{sun2022style}). (b) Introducing artistic concepts as controllable parameters (e.g., Concrete and Rough Oil Paintings~\cite{tong2022im2oil}). (c) Mimicking artistic genres by representing them as technical problems (e.g., Multi-View Optical Illusions~\cite{geng2023visual}). (d) Designing frameworks to align different application scenarios and artistic requirements (e.g., Non-typical Aspect Image Generation for Visual Storytelling~\cite{wang2023magicscroll}). (e) Surpassing 2D painting by achieving higher-dimensional controls (e.g., Multi-View Wire Art~\cite{qu2023wired}). (f) Applying generative techniques to facilitate application scenarios  (e.g., typography design~\cite{tanveer2023ds}).}
%     \label{fig:visual_art}
% \end{figure}

\subsection{Research Goals and Questions}
Following the previous discussion and prior work, we summarize two research goals of our paper:
\begin{enumerate}[label=\textbf{G\arabic*}]
    \item \textbf{Analyze how diffusion-based methods have facilitated and transformed visual art creation.} How are diffusion-based generative systems and models used for different Visual Art applications? 
    \item \textbf{Provide frameworks, trends, and inspirations for future research in relevant fields.} How may human and generative AI inspire each other in Diffusion-Based Visual Art Creation? 
\end{enumerate}

Based on the two research goals, we further propose four research questions as the basis of this survey.

\begin{enumerate}[label=\textbf{Q\arabic*}]
    \item \textbf{What are the most attended topics in diffusion-based Visual Art Creation?} This is the basic step to identify hot issues and construct a consistent framework. The question also concerns contrasts between diffusion-based and non-diffusion-based methods, and the temporal features and evolution of this field.
    \item \textbf{What are current research problems/needs/requirements in diffusion-based Visual Art creation?} This question ranges from an artistic/user perspective to a technical/designer perspective. In the following sections, we further break it down to artistic requirements and technical problems, featured by application scenarios, data modalities, and generative tasks, and attempt to establish connections between them.
    \item \textbf{What are the methods applied in diffusion-based Visual Art creation?} For each technical problem, we focus on diffusion-based method design according to its modalities and tasks. As DDPM, DDIM, and their extensions follow similar model structures, we can further categorize and organize the methods based on the unified structure of an extended diffusion model.
    \item \textbf{What are the frontiers, trends, and future works?} We are interested in the following questions: Are there any further problems to solve? How may we leverage the development of a diffusion model and its application in relevant fields to cope with the problems?
\end{enumerate}

\section{Findings}
In this section, we aim to fulfill \textbf{G1} by answering questions \textbf{Q1--Q3}.

\subsection{Structural Analysis and Framework Construction}
\label{sec: framework}
\subsubsection{Data Classification.} 
\begin{figure}
    \centering
    \includegraphics[width=\textwidth]{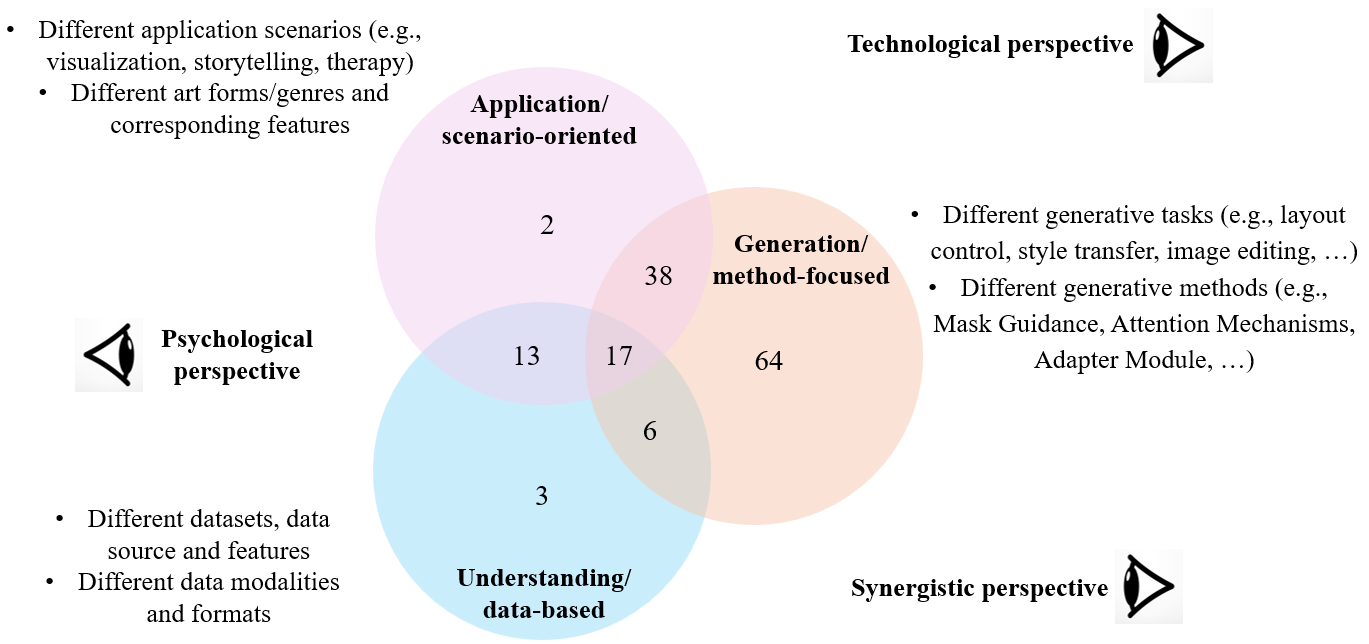}
    \caption{Venn Chart for Topics in Visual Art Creation. The chart is summarized from data distribution and annotations in our dataset. This framework is used to categorize and distinguish the blueprints of relevant research (Sec.~\ref{sec: framework}) and to analyze the development and current state of this field (Sec.~\ref{sec: temporal}). In Sec.~\ref{sec: discussion}, we further provide technological, synergistic, and application perspectives as extensions of these three categories for development trends and future work.}
    \label{fig:venn}
\end{figure}

We focus on the first question: What are the currently most attended topics in diffusion-based Visual Art Creation? (\textbf{Q1}) We first summarized different paper codes proposed in Sec.~\ref{sec: coding} along with the index terms of each selected paper. Among them, a major part is closely related to method design, while others concern data, modalities, artistic genres, and application scenarios. We found that these terms basically form three categories and thus applied a Venn Chart to characterize different works, as shown in Fig.~\ref{fig:venn}. The three categories include:
\begin{itemize}
    \item \textbf{Application.} Different application scenarios (e.g., different art genres~\cite{geng2023visual, qu2023wired}, visualization~\cite{zeng2024intenttuner, xiao2024typedance}) 
    \item \textbf{Understanding.} Different data forms and corresponding modalities (e.g., image series~\cite{song2020character, suh2022codetoon, braude2022ordered}, 3D scenes~\cite{dong2024interactive3d, zhang2024coarf, haque2023instruct}). From an artistic perspective, the first two categories characterize different art forms/genres with corresponding features.
    \item \textbf{Generation.} Different generative tasks (e.g., style control~\cite{liang2023adversarial, wu2023not}, style transfer~\cite{zhang2024artbank, zhang2024towards}, image editing~\cite{hertz2022prompt, biner2024sonicdiffusion}) and Different generative methods (e.g., ControlNet~\cite{xue2024strictly, lukovnikov2024layout}, Textual Inversion~\cite{ahn2024dreamstyler, zhang2023inversion}, LoRA~\cite{shah2023ziplora, chen2024id})
\end{itemize}

With such a categorization method, we can approach the dataset from different perspectives and identify corresponding hot topics. As shown in Fig.~\ref{fig:venn}, most of the selected works lie in four subsets of the seven areas, including:
\begin{itemize}
    \item \textbf{Generation (125).} Generation and editing with controllable style, subject, and layout (e.g., personalization~\cite{wu2024u, nam2024dreammatcher}, stylization~\cite{tanveer2023ds, wang2023styleadapter}, layout control~\cite{yamada2024glod, couairon2023zero})
    \item \textbf{Generation $\cap$ Application (55).} Application-oriented generation (e.g., art therapy~\cite{zubala2021art}, visual art education~\cite{dehouche2023s}, computational arts metaverse~\cite{lee2021creators})
    \item \textbf{Application $\cap$ Understanding (30).} Mimicking specific historical context/ artistic genre (e.g., Chinese calligraphy~\cite{wang2023naturality}, Indian Art~\cite{kolay2016cultural}, St. Paul painting~\cite{d2021combined})
    \item \textbf{Generation $\cap$ Understanding (23).} Analysis and understanding by generation (e.g., reflection on the essence of art and creativity~\cite{ye2023everyone}, exploring new concepts and possibilities~\cite{song2023expanded})
\end{itemize}

\subsubsection{Framework Construction.} 
\begin{figure}
    \centering
    \includegraphics[width=\textwidth]{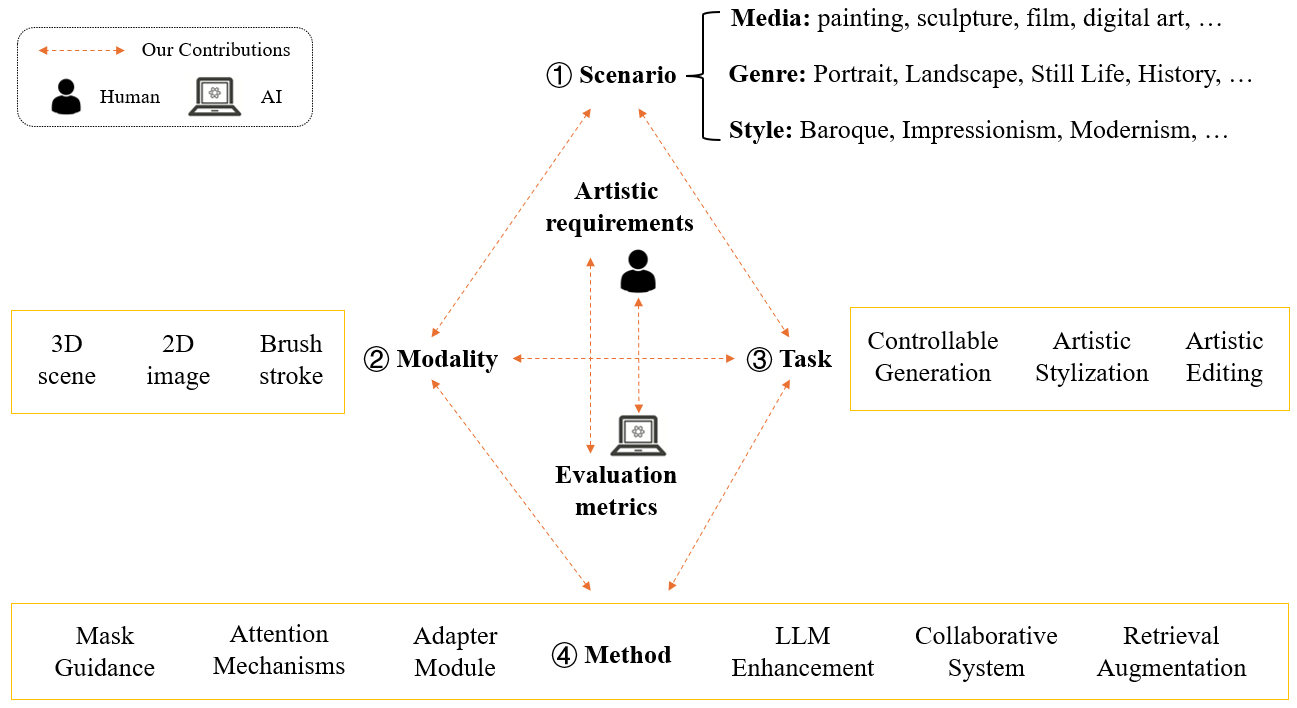}
    \caption{An Overall Framework for Diffusion-Based Visual Art Creation. The main contributions of this paper lie in establishing the connections between scenario, modality, task, and method, as well as outlining a general roadmap from artistic requirements (human perspective) to technical problems (AI perspective). This framework is then used to analyze each individual paper in our dataset (Sec.~\ref{sec: analysis} and Sec.~\ref{sec: method_design}).}
    \label{fig:framework}
\end{figure}

Since most of the work is concentrated at the pole of generation, we dived into the generative part and paid more attention to the intersection areas. We found that 1) Research in Diffusion-Based Visual Art Creation is typically characterized by different artistic scenarios and technical methods. 2) the artistic requirements and technical problems are basically connected by specifying data modality and extracting generative tasks. As a result, we summarized a new framework that can better characterize the current research paradigm (Fig.~\ref{fig:framework}). 

Based on the framework, we can further break down \textbf{Q1} into a series of consequential questions and approach a generative problem from different perspectives:

\begin{itemize}
    \item \textbf{Scenario.} What are the common features and requirements of different artistic scenarios? 
    \item \textbf{Modality.} What are the data modalities applied, including training dataset, input, conditions, and output?
    \item \textbf{Task.} What are popular research problems in generating Visual Arts, including their technical statements and classification?
    \item \textbf{Method.} What are the methods used to augment and adapt diffusion models?
\end{itemize}

In the following sections, we will refer to this structure to analyze the relationships represented by each red dashed line. 

\subsection{Temporal Analysis and Trend Detection.}
\label{sec: temporal}
In this part, we investigate how the number of publications, categories, and keywords in different dimensions evolve over time in our dataset. We specifically focus on the difference between pre-diffusion and post-diffusion eras.

\subsubsection{Data Distribution.} 

\begin{figure}
    \centering
    \includegraphics[width=\textwidth]{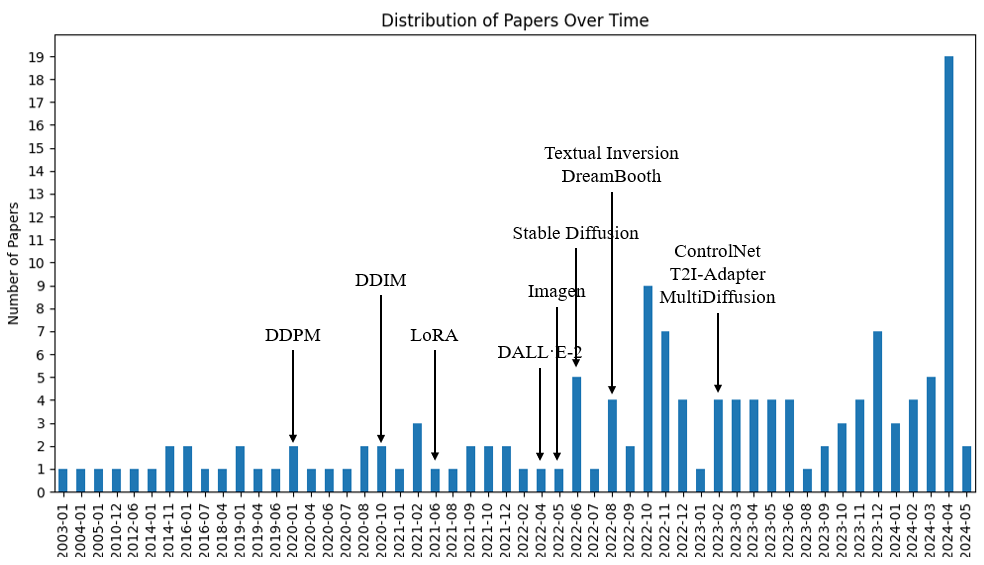}
    \caption{Temporal Distribution of the Number of Papers in Our Selected Dataset. We also labeled the timestamps when major models are proposed~\cite{ho2020denoising, song2020denoising, hu2021lora, ramesh2022hierarchical, saharia2022photorealistic, rombach2022high, gal2022image, ruiz2023dreambooth, zhang2023adding, mou2024t2i, bar2023multidiffusion}.}
    \label{fig:data_distribution}
\end{figure}

\begin{figure}
    \centering
    \includegraphics[width=\textwidth]{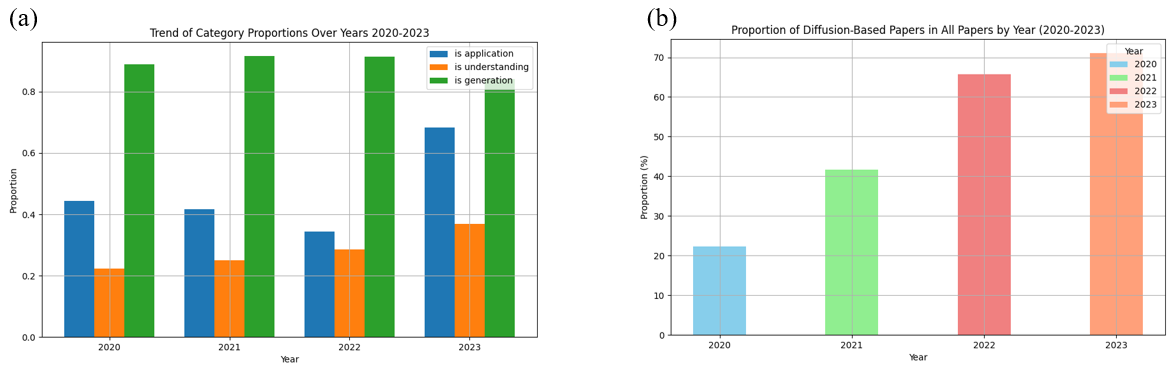}
    \caption{Temporal Statistics of Research Categories in Our Selected Dataset. (a) Yearly Trends of the Proportions of Papers in Three Categories. (b) Diffusion-Based Papers after the Diffusion Era (2020-2023).}
    \label{fig:yearly_trends}
\end{figure}

Fig.~\ref{fig:data_distribution} displays the temporal distribution of our dataset, including the time when major diffusion-based models are proposed. According to the figure, most of the work is published after Jun. 2022, especially after Feb. 2023, when a series of landmark methods were introduced. We also calculate the proportions of different categories of work and demonstrate the result in Fig.~\ref{fig:yearly_trends}. We found that while the proportion of generation research remains essentially the same, understanding research keeps growing in the four years of 2020-2023, and application research witnessed a surge in 2023. On the other hand, the proportion of diffusion-based methods increased steadily, from around 20\% in 2020 to over 70\% in 2023.

\subsubsection{Topic Evolution.} 
\begin{figure}
    \centering
    \includegraphics[width=\textwidth]{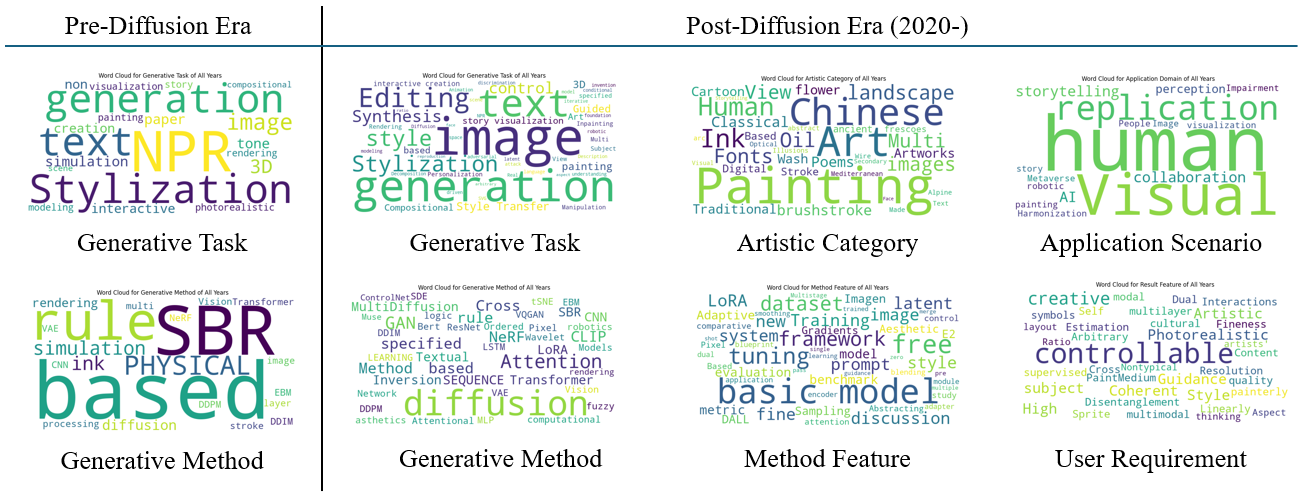}
    \caption{Multi-Dimensional Word Clouds in Our Selected Dataset. By coding each paper and collecting keywords from different perspectives, we derive multiple word clouds of our dataset for both before and after the diffusion era (2020).}
    \label{fig:word_cloud}
\end{figure}

In examining the dataset, we noticed that most methods and some tasks change and upgrade over time, while application scenarios are generally constant. We thus provide a Word Cloud Chart (Fig.~\ref{fig:word_cloud}) to compare methods and tasks in pre-diffusion and post-diffusion eras and illustrate the main application scenarios, artistic categories, method features, and user requirements. According to the figure, most generative tasks stay the same, while some traditional tasks (e.g., NPR) are less mentioned. However, generative methods have undergone a major shift, from SBR (stroke-based rendering), rule-based generation, and physical-based simulation, to a series of modifications based on diffusion models. 

\subsubsection{Qualitative Comparison.} From a microscopic perspective, we are also interested in how the development of diffusion-based models introduces new methods for solving traditional problems. We thus selected five artistic genres or scenarios, including robotic painting, Chinese landscape painting, ink painting, story visualization, and artistic font synthesis, to compare different approaches for similar problems and tasks.
\label{sec: comparison}

\begin{figure}
    \centering
    \includegraphics[width=0.7\textwidth]{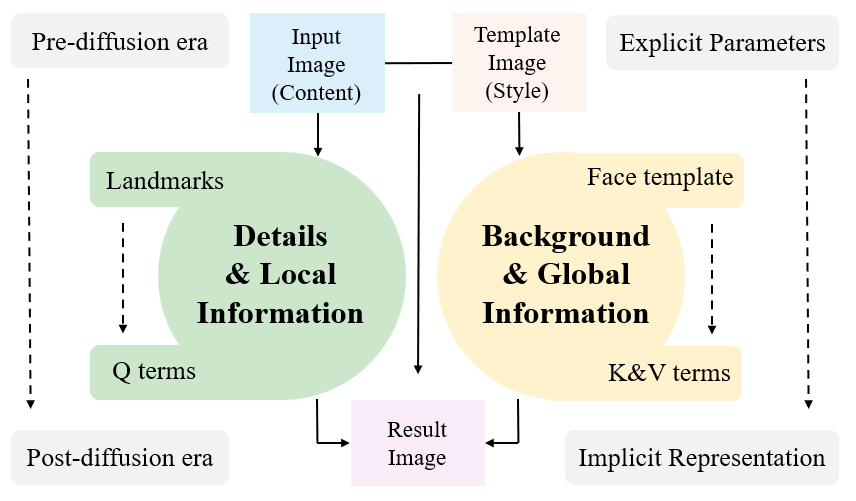}
    \caption{Comparison Between Different Methods on Similar Tasks Before and After the Diffusion Era. We select the task of portrait stylization as an example. We compare traditional methods based on image blending~\cite{chiang2018generation} and the model structure of the newly-proposed Portrait Diffusion~\cite{liu2023portrait}. Based on them, we abstract a unified paradigm for portrait stylization.}
    \label{fig:comparison} 
\end{figure}

Fig.~\ref{fig:comparison} displays an example to compare different methods for similar tasks (portrait stylization) before and after the Diffusion Era. According to the similarities between the two workflows, we derived a unified, model-agnostic structure for solving the problem. As shown in Fig.~\ref{fig:comparison}, in solving the task of face stylization (stylized portrait generation), both methods generate the new image based on the previous two images. Meanwhile, they both refer to the input image (content reference) for details and local information, and the template image (style reference) for background and global information in the generative process. When traditional tasks meet with new methods, such frameworks provide an interesting perspective to capture the embedded human expertise. 
%主要从结果质量出发，参考模型复杂度，计算时间等
On the other hand, the two approaches display multiple differences in result quality, model complexity, computational cost, etc. Based on multiple pairs of selected work in our dataset, we summarize the following trends:
\begin{itemize}
    \item \textbf{Input Format:} from images only to images and masks as conditional input (controllability↑)
    \item \textbf{Dataset:} from fixed database to arbitrary image (generalizability↑)
    \item \textbf{Generative Process:} from explicit/pixel manipulation to implicit/latent manipulation (Complexity↑)
    \item \textbf{Method Category:} from traditional rule-based image processing to diffusion-based image stylization (Computational cost/Time consumption↑)
\end{itemize}

\subsubsection{A Brief Summary and Outlook.}

In the previous section, we have compared methods before and after the Diffusion Era, to borrow frameworks and ideas from the Pre-Diffusion Era and inspire new method design. Here we are further interested in identifying research gaps from temporal trends and task-method relationships in Diffusion-Based Visual Art Creation.

% table
\begin{table}[h]
\centering
\caption{Top Growing Keywords in Method Features and User Requirements by Year. }
\label{tab: comparison}
\begin{tabular}{ccc}
\toprule
Year & Method Features & User Requirements \\
\midrule
2021 & Basic Model, Dataset, Metric, Evaluation & (No specific requirements) \\
2022 & Framework, Adaptive, Sampling, Fine-tuning & Photorealistic, Multilayer, Artistic, Creative, Coherent \\
2023 & Tuning-free, Training-free, System, Prompt & Controllable, Subject, Disentanglement, Interaction, Painterly \\
2024 & Inversion, Dilation, Layer-aware, Step-aware & Personalization, Composition, Visualization, Concept, Context \\
\bottomrule
\end{tabular}
\end{table}

In Table~\ref{tab: comparison}, we present the top growing keywords in method features and user requirements for the post-diffusion era. We first calculate keyword frequencies for each year from 2020 to 2024 (up until May 2024). We then calculate the difference in each word's frequency compared to the previous year. The words with the highest frequency growth are identified as 'Top Growing Key Words'. Combine Table~\ref{tab: comparison} with Fig.~\ref{fig:word_cloud}, we identify some major trends in Diffusion-Based Visual Art Creation:
\begin{itemize}
    \item \textbf{Technically}, the research type developed from a basic model to a generative framework to an interactive system. Researchers' design focus also shifted from developing benchmarks (dataset, metric, evaluation) to introducing generative methods (sampling, inversion, dilation), with a general trend to simplify the generative process (tuning-free and training-free).
    \item \textbf{Artistically}, user requirements are diverging from higher quality (photorealistic, artistic, coherent) to multiple diversified needs (controllable, composition, visualization), and research focus has shifted from the generated visual content (multilayer, coherent) to creative subject (personalization, concept, context).   The most notable requirement is \emph{creative}, which emerged two years ago but has not been well resolved until now (Sec.~\ref{sec: creativity}).
    \item \textbf{Interdisciplinarily}, the keywords also manifested more collaborations between human creators and AI models. On the one hand, experts introduced principles in the diffusion process (e.g., step-aware) and concepts from artistic areas (e.g., layer-aware), to boost controllability and performance. On the other hand, researchers ventured into understanding implicit latent (e.g., disentanglement), adapting the system to user inputs (e.g., prompt), and catering the diffusion model to the human thinking process (e.g., interaction). 
\end{itemize}

In Sec.~\ref{sec: discussion}, we will go into detail to discuss the trends and future outlook from multiple perspectives.

\subsection{From Artistic Requirements to Technical Problems}
\label{sec: analysis}

In this section, we focus on the upper half of the Rhombus framework (Fig.~\ref{fig:framework}), to summarize current research problems/needs/requirements in Diffusion-Based Visual Art creation (\textbf{Q2}). Specifically, we start from application scenarios and artistic genres, analyze their corresponding data modality and generative task, and dive into their key requirements/goals and computational statements. By doing this, we aim to fill in the three connections and bridge the gap between artistic requirements and technical problems. 

\subsubsection{Application Domain and Artistic Category.}

Among our 143 selected papers, 70 are coded as application/scenario-oriented. Within this subset, 55 papers focus on specific artistic categories (e.g., traditional paintings, human portraits, and specific art genres), and 17 focus on relevant domains (e.g., story visualization, replication prevention, human-AI collaboration). We summarize representative work in different application scenarios, focusing on how they formulate and tackle the domain issues.

% 列举不同artistic category和application domain的工作
The first series of works view visual art (or digital art, fine art) as a general category. Abrahamsen et al.~\cite{Abrahamsen2023InventingAS} introduce innovative methods to invent art styles using models trained exclusively on natural images, thereby circumventing the issue of plagiarism in human art styles. Their approach leverages the inductive bias of artistic media for creative expression, harnessing abstraction through reconstruction loss and inspiration from additional natural images to forge new styles. This holds the promise of ethical generative AI use in art without infringing upon human creators' originality.
In a similar vein, Zhang et al.~\cite{zhang2024artbank} address the limitations of existing artistic style transfer methods, which either fail to produce highly realistic images or struggle with content preservation, by proposing ArtBank. This novel framework, underpinned by a Pre-trained Diffusion Model and an Implicit Style Prompt Bank (ISPB), adeptly generates lifelike stylized images while maintaining the content's integrity. The added Spatial-Statistical-based self-attention Module (SSAM) further refines training efficiency, with their method surpassing contemporary artistic style transfer techniques in both qualitative and quantitative evaluations.
Meanwhile, Qiao et al.~\cite{qiao2022initial} explore the use of image prompts in conjunction with text prompts to enhance subject representation in multimodal AI-generated art. Their annotation experiment reveals that initial images significantly improve subject depiction, particularly for concrete singular subjects, with icons and photos fostering high-quality, aesthetically varied generations. They provide valuable design guidelines for leveraging initial images in AI art creation.
Furthermore, Huang et al.~\cite{huang2022draw} present the multimodal guided artwork diffusion (MGAD) model, a novel approach to digital art synthesis that leverages multimodal prompts to direct a classifier-free diffusion model, thereby achieving greater expressiveness and result diversity. The integration of the CLIP model unifies text and image modalities, with substantial experimental evidence endorsing the efficacy of the diffusion model coupled with multimodal guidance.
Lastly, Liao et al.~\cite{liao2022artbench} contribute to the field by introducing ArtBench-10, a class-balanced, high-grade dataset for benchmarking artwork generation. It stands out with its clean annotations, high-quality images, and standardized dataset creation process, addressing the skewed class distributions prevalent in prior artwork datasets. Available in multiple resolutions and formatted for seamless integration with prevalent machine learning frameworks, ArtBench-10 facilitates comprehensive benchmarking experiments and in-depth analyses to propel generative model research forward.
Collectively, these works illustrate the dynamic intersection of AI and art, where innovative methodologies and datasets are expanding the frontiers of artistic creation, opening avenues for novel styles, ethical considerations, and enhanced representation in the digital art sphere.

The second series of works focus on specific artistic genres or historical contexts, among which traditional Chinese painting is most frequently visited. Wang et al.~\cite{wang2023cclap} introduce CCLAP, a pioneering method for controllable Chinese landscape painting generation. By leveraging a Latent Diffusion Model, CCLAP consists of a content generator and style aggregator that together produce paintings with specified content and style, evidenced by both qualitative and quantitative results that showcase the model's artful composition capabilities. A dedicated dataset, CLAP, has been developed to evaluate the model comprehensively, and the code has been made accessible for broader use.
Addressing the issue of low-resolution images in the digital preservation of Chinese landscape paintings, Lyu et al.~\cite{lyu2024diffusion} propose the diffusion probabilistic model CLDiff. It employs iterative refinement steps akin to the Langevin dynamic process to transform Gaussian noise into high-quality, ink-textured super-resolution images, while a novel attention module enhances the U-Net architecture's generative power.
Fu et al.~\cite{fu2021multi} tackle the challenge of generating traditional Chinese flower paintings with various styles such as line drawing, meticulous, and ink through a deep learning approach. Their Flower-Generative Adversarial Network framework, bolstered by attention-guided generators and discriminators, facilitates style transfer and overcomes common artifacts and blurs. A new loss function, Multi-Scale Structural Similarity, is introduced to enforce structural preservation, resulting in higher quality multi-style Chinese art paintings.
From the perspective of generative teaching aids, Wang et al.~\cite{wang2024intelligent} present "Intelligent-paint," a method for generating the painting process of Chinese artworks. Using a Vision Transformer-based generator and an adversarial learning approach, this method emphasizes the unique characteristics of Chinese painting, such as void and brush strokes, employing loss constraints to align with traditional techniques. The coherence of the generated painting sequences with real painting processes is further validated by expert evaluations, making it a valuable tool for beginners learning Chinese painting.
Finally, Li et al.~\cite{li2021paint4poem} introduce the novel task of artistically visualizing classical Chinese poems. For this, they construct the Paint4Poem dataset, comprising high-quality poem-painting pairs and a larger collection to assist in training poem-to-painting generation models. Despite the models' capabilities in capturing pictorial quality and style, reflecting poem semantics remains a challenge. Paint4Poem opens many research avenues, such as transfer learning and text-to-image generation for low-resource data, enriching the intersection of literature and visual art.
These works collectively highlight the potential of diffusion-based techniques in enriching the field of traditional Chinese painting, offering advanced tools for both creation and restoration and enhancing the educational process for aspiring artists.

With the development of diffusion-based generative methods, the application scenario has expanded to cover a wide range of artistic categories, including human images, portraits, fonts, and more.
Ju et al.~\cite{ju2023human} have crafted the Human-Art dataset to bridge the gap between natural and artificial human representations. Spanning natural and artificial scenes, this dataset is comprehensive, covering 2D and 3D instances, and is poised to enable advancements in various computer vision tasks such as human detection, pose estimation, image generation, and motion transfer.
Liu et al.~\cite{liu2023portrait} present Portrait Diffusion, a training-free face stylization framework that utilizes text-to-image diffusion models for detailed style transformation. This novel framework integrates content and style images into latent codes, which are then delicately blended using Style Attention Control, yielding precise face stylization. The innovative Chain-of-Painting method allows for gradual redrawing of images from coarse to fine details.
In the realm of secondary painting for artistic productions like comics and animation, Ai et al.~\cite{ai2023stable} introduce Stable Diffusion Reference Only, a method that accelerates the process with a dual-conditioning approach using image prompts and blueprint images for precise control. This self-supervised model integrates seamlessly with the original UNet architecture, enhancing efficiency and controllability without the need for complex training methods.
Wang et al.~\cite{wang2023magicscroll} tackle the challenge of creating nontypical aspect-ratio images with MagicScroll, a diffusion-based image generation framework. It addresses issues of content repetition and style inconsistency by allowing fine-grained control of the creative process across object, scene, and background levels. This model is benchmarked against mediums like paintings, comics, and cinema, demonstrating its potential in visual storytelling.
Lastly, Tanveer et al.~\cite{tanveer2023ds} introduce DS-Fusion, a method for generating artistic typography that balances stylization with legibility. Utilizing large language models and an unsupervised generative model with a diffusion model backbone, it creates typographies that visually convey semantics while remaining coherent. DS-Fusion is validated through user studies and stands out against prominent baselines and artist-crafted typographies.
Together, these advancements signify a major leap in the application of diffusion-based methods to a myriad of artistic categories. By encompassing human-centric datasets, training-free frameworks, speed-enhancing models for artists, tools for visual storytelling, and typography generation techniques, the scope of AI in art creation is being pushed to new, previously unimagined heights.

\subsubsection{Representing Scenarios as Modalities and Tasks.}

Next, we attempt to structure different application scenarios by their corresponding data modalities and generative tasks. In this way, we aim to approach the embedded technical problems and establish alignment between artistic requirements and technical problems.

Following the common practice in AIGC, we first categorize artistic scenarios by different data modalities:
\begin{itemize}
    \item \textbf{Thread/Brushstroke.} The first series of work focus on brush stroke generation. The problem has been long-studied and technically attended since around 2000 and can be well solved by traditional rendering and rule-based methods, with little involvement of diffusion-based models~\cite{lee2005real, yang2019easy, nakano2019neural, bidgoli2020artistic, fang2018automatic}.
    \item \textbf{2D Pixels/Image.} Among all modalities, 2D images are the most common representation in visual art, and thus a great bunch of work adopts it as (one of) the target representations~\cite{lin2023calligraphy, he2023scalecrafter, zhu2024multibooth, bai2024real, cong2024automatic}.
    \item \textbf{Image Series/Video.} Typically considered as a temporal extension or duplication of a single image, image series and videos are common in certain scenarios such as storytelling and animation~\cite{song2020character, braude2022ordered, suh2022codetoon, wang2023simonstown, gong2023interactive}. 
    \item \textbf{3D Model/Scene.} Some art forms are based on spatial expression, and the field of 3D generation is also extending its artistic perspective, and thus 3D visual art creation is growing rapidly~\cite{dong2024interactive3d, qu2023wired, zhang2022arf, wang2024headevolver, haque2023instruct}. 
    \item \textbf{Others.} Other artistic genres are commonly believed to possess certain modality features. For example, sketch share both a raster and a vector representation, thus inspiring researchers to take different generative approaches~\cite{li2024inverse, wang2022learning, wang2022enhancing, wang2021tracing, Ciallo2024}.
\end{itemize}

Next, we summarize typical tasks in Diffusion-Based Visual Art Creation: 
\begin{itemize}
    \item \textbf{Quality Enhancement.} As the baseline task in content generation and the basic requirement in visual art creation, the generated content should possess higher resolution and better quality. This is commonly realized by aesthetic training data, advanced model structure, more parameters, and result optimization designs. In the post-diffusion era, these methods are integrated into training foundation models~\cite{he2023scalecrafter, nichol2021glide, chen2023pixart, zhang2024towards, rombach2022high}.
    \item \textbf{Controllable Generation.} The requirement emerges from artists' need to precisely control each perspective of their generated results, including context, subject, content, and style. Researchers adapt diversified ways, including additional information encoding, cross-attention mechanism, and retrieval augmentation, to support different forms of conditions~\cite{zhang2023adding, li2023gligen, ye2023ip, huang2024diffstyler, zhao2024uni}.
    \item \textbf{Content Editing and Stylization.} This task is seen in various scenarios such as iterative generation, collaborative creation, and image inpainting. Following the understanding of high-level concept and low-level style in deep latent structure, experts are also working on decoupling the two aspects, to improve the performance of diffusion-based models on style transfer, style control, style inversion, etc~\cite{hertz2022prompt, lu2023painterly, brack2022stable, Abrahamsen2023InventingAS, kawar2023imagic}.
    \item \textbf{Specialized Tasks.} According to different visual art scenarios and inspired by human concepts, experts summarized and proposed new tasks including compositional generation (e.g., concept, layout, layer) and latent manipulation. Still, more research is application-oriented, designed, and optimized for specific data types (e.g., human portrait) or specific scenarios (e.g., multiview art)~\cite{wang2023magicscroll, wu2023not, zhou2023clip, chefer2023attend, liu2022compositional}.
\end{itemize}

\subsubsection{From Artistic Goals to Evaluation Metrics.} In Diffusion-Based Visual Art Creation, artistic goals drive the development of generative tasks, and the success of these tasks is measured using specific evaluation metrics. In Table~\ref{tab:evaluation}, we summarize common artistic goals and list several evaluation metrics as an example:

\begin{itemize}
\item \textbf{Controllability.} Achieving precise control over generated outcomes is measured by metrics that evaluate the adherence to user-specified prompts and directions.
\begin{itemize}
\item CLIP Score: Assesses alignment between text prompts and generated images using CLIP (Contrastive Language-Image Pretraining) embeddings~\cite{radford2021learning}.
\item CLIP Directional Similarity: Measures the semantic similarity between changes in text prompts and corresponding changes in generated images~\cite{patashnik2021styleclip}.
\end{itemize}
\item \textbf{Visual Quality.} The quality of generated art is quantified by subjective and objective metrics that reflect the aesthetic and technical excellence of the artwork.
\begin{itemize}
\item User studies: Subjective evaluations where users rate the visual appeal and aesthetic qualities of generated content~\cite{wang2023imagen}.
\item LAION-AI Aesthetics: A metric that uses a dataset from LAION-AI to objectively evaluate the aesthetic aspects of generated images, such as harmony, balance, and composition~\cite{schuhmann2021laion}.
\end{itemize}
\item \textbf{Fidelity.} The fidelity of the generated content to the target data distribution is gauged using metrics that compare the statistical properties of generated and real artwork.
\begin{itemize}
\item FID (Fréchet Inception Distance)~\cite{heusel2017gans}: Quantifies the distance between feature distributions of generated and real images to assess the realism and diversity of the content~\cite{liu2023instaflow}.
\item IS (Inception Score)~\cite{salimans2016improved}: Measures the clarity and variety of generated images based on the Inception network's confidence in classifying the content and the diversity across the dataset~\cite{dhariwal2021diffusion}.
\end{itemize}
\item \textbf{Interpretability.} For the goal of interpretability, metrics assess how well we can understand and manipulate the generative model's inner workings.
\begin{itemize}
\item Disentanglement metrics: Utilize methods such as the $\beta$-VAE metric~\cite{higgins2017beta} to quantify the independence of different factors in the latent variables~\cite{jin2024closed}.
\item Feature attribution: Employ techniques such as SHAP (SHapley Additive exPlanations)~\cite{lundberg2017unified} to determine which features or latent variables have the greatest impact on the characteristics of the generated content ~\cite{kowalek2022boosting}.
\end{itemize}
\end{itemize}

% table
\begin{table}[h]
\centering
\caption{Correspondence between Artistic Goals and Evaluation Metrics.}
\begin{tabular}{cc}
\toprule
Artistic Goal & Example Evaluation Metric \\
\midrule
Controllability &  CLIP Score~\cite{radford2021learning}, CLIP Directional Similarity~\cite{patashnik2021styleclip} \\
Visual Quality & User studies,  LAION-AI Aesthetics~\cite{schuhmann2021laion} \\
Fidelity & Fréchet Inception Distance~\cite{heusel2017gans}, Inception Score~\cite{salimans2016improved} \\
Interpretability & Disentanglement metrics, feature attribution \\
\bottomrule
\end{tabular}
\label{tab:evaluation}
\end{table}

\subsection{Design and Application of Diffusion-Based Methods}
\label{sec: method_design}

In the previous discussion, we gradually shifted from an artistic/user perspective to a technical/designer perspective. In this part, we focus on the lower half of the Rhombus framework (Fig.~\ref{fig:framework}), to summarize specific methods applied in Diffusion-Based Visual Art Creation (\textbf{Q3}).

\subsubsection{From Generative Tasks to Method Design.}

Based on the previously summarized generative tasks, we first categorize representative diffusion-based methods applied to solve each problem. We specifically focus on controllable generation, content editing, and stylization which together take up more than 80\% of research focuses in generative/method-based research.

\textbf{Controllable generation.} In the realm of controllable generation, various studies have presented innovative approaches to guide diffusion models effectively.
The work by Choi et al.~\cite{choi2021ilvr} introduces Iterative Latent Variable Refinement (ILVR), which conditions denoising diffusion probabilistic models (DDPM) using a reference image. The ILVR method directs a single DDPM to generate images with various attributes informed by the reference, enhancing the controllability and quality of generated images across multiple tasks like multi-domain image translation and image editing.
Gal et al.~\cite{gal2022image} propose a method that personalizes text-to-image generation by learning new "words" to represent user-provided concepts. This approach, named Textual Inversion, adapts a frozen text-to-image model to generate images of unique concepts. By embedding these unique "words" into natural language sentences, users have the creative freedom to guide the AI in generating personalized images.
In another breakthrough, Zhang et al.~\cite{zhang2023adding} present ControlNet, an architecture that adds spatial conditioning controls to pre-trained text-to-image diffusion models. ControlNet takes advantage of "zero convolutions" and existing deep encoding layers from large models, allowing the fine-tuning of conditional controls like edges and segmentation with robust training across different dataset sizes.
Building further on control mechanisms, Zhao et al.~\cite{zhao2024uni} introduce Uni-ControlNet, a unified framework that enables the simultaneous use of multiple control modes, both local and global, without the need for extensive training from scratch. The framework's unique adapter design ensures cost-effective and composable control, enhancing both controllability and generation quality.
Finally, Ruiz et al.~\cite{ruiz2023dreambooth} present DreamBooth, a fine-tuning approach that personalizes text-to-image diffusion models to generate novel renditions of subjects in varying contexts using a small reference set. This method, empowered by a class-specific prior preservation loss, maintains the subject's defining features across different scenes, opening the door to new applications like subject recontextualization and artistic rendering.
These studies collectively illustrate the evolving landscape of design and application within diffusion-based methods. They highlight the progress from generative tasks to refined method design and the ongoing pursuit of enhanced controllability in image generation.

\textbf{Content Editing.} The design and application of diffusion-based methods have paved the way for breakthroughs in content editing, offering enhanced photorealism and greater control in the text-guided synthesis and manipulation of images.
Nichol et al.~\cite{nichol2021glide} delve into text-conditional image generation using diffusion models, contrasting CLIP guidance with classifier-free guidance. The latter is favored for producing realistic images that closely align with human expectations. Their 3.5 billion parameter model outperforms DALL-E in human evaluations, and further demonstrates its flexibility in image inpainting, facilitating text-driven editing capabilities.
Hertz et al.~\cite{hertz2022prompt} introduce an intuitive image editing framework, where modifications are steered solely by textual prompts, bypassing the need for spatial masks. Their analysis highlights the crucial role of cross-attention layers in mapping text to image layout, enabling precise control over local and global edits while preserving fidelity to the original content.
Kumari et al.~\cite{kumari2023multi} propose an efficient approach for incorporating user-defined concepts into text-to-image diffusion models, Custom Diffusion. By optimizing a subset of parameters, the method allows for rapid adaptation to new concepts and the combination of multiple concepts, yielding high-quality images that outperform existing methods in both efficiency and effectiveness.
Brooks et al.~\cite{brooks2023instructpix2pix} present InstructPix2Pix, a conditional diffusion model trained on a dataset generated by combining the expertise of GPT-3 and Stable Diffusion. This model can interpret human-written instructions to edit images accurately, operating swiftly without needing per-example fine-tuning, showcasing its proficiency across a wide array of editing tasks.
Lastly, Parmar et al.~\cite{parmar2023zero} tackle the challenge of content preservation in image-to-image translation with pix2pix-zero. Through the discovery of editing directions in text embedding space and cross-attention guidance, their method ensures the input image's content remains intact. They further streamline the process with a distilled conditional GAN, achieving superior performance in both real and synthetic image editing without necessitating additional training.
Collectively, these advancements in diffusion-based methods signify a transformative period in content editing, where the synthesis of images is becoming increasingly controllable, customizable, and responsive to textual nuance, greatly expanding the potential for creative expression and practical applications.

\textbf{Stylization.} Recent advancements in diffusion-based methods have significantly enhanced the stylization capabilities in the domain of generative AI, enabling more intuitive and precise artistic expression.
Zhang et al.~\cite{zhang2023inversion} propose an inversion-based style transfer technique that captures the artistic style directly from a single painting, circumventing the need for complex textual descriptions. This method, named InST, efficiently captures the essence of a painting's style through a learnable textual description and applies it to guide the synthesis process, thus achieving high-quality style transfer across diverse artistic works.
Huang et al.~\cite{huang2024diffstyler} present DiffStyler, a novel architecture that leverages dual diffusion processes to control the balance between content and style during text-driven image stylization. By integrating cross-modal style information as guidance and proposing a content image-based learnable noise, DiffStyler ensures that the structural integrity of the content image is maintained while achieving a compelling style transformation.
In the realm of artistic image synthesis, Ahn et al.~\cite{ahn2024dreamstyler} propose DreamStyler, a framework that optimizes multi-stage textual embedding with context-aware text prompts. DreamStyler excels at both text-to-image synthesis and style transfer, providing the flexibility to adapt to various style references and producing images that exhibit high-quality and unique artistic traits.
Sohn et al.~\cite{sohn2024styledrop} develop StyleDrop, a method designed to synthesize images that adhere closely to a specific style using a text-to-image model. StyleDrop stands out for its ability to capture intricate style nuances with minimal parameter fine-tuning. It demonstrates impressive results even when provided with a single image, effectively synthesizing styles across different patterns, textures, and materials.
Together, these methodologies exemplify the ongoing innovation in the field of image stylization through diffusion-based methods. They afford users an unprecedented level of control and flexibility in generating and editing images, breaking new ground in the creation of stylized artistic content. These tools not only facilitate the expression of visual art but also promise to expand the possibilities for personalized and creative digital media.

\textbf{Quality Enhancement.}
The exploration of diffusion-based methods has led to significant enhancements in the quality of text-to-image synthesis, pushing the boundaries of resolution, fidelity, and customization.
Balaji et al.~\cite{balaji2022ediff} propose eDiff-I, an ensemble of text-to-image diffusion models that specialize in different stages of the image synthesis process. This approach results in images that better align with the input text while maintaining visual quality. The models use various embeddings for conditioning and introduce a "paint-with-words" feature, which allows users to control the output by applying words to specific areas of an image canvas, providing a more intuitive way to craft images.
Chang et al.~\cite{chang2023muse} introduce Muse, a Transformer model that surpasses diffusion and autoregressive models in efficiency. Muse achieves state-of-the-art performance with a masked modeling task on discrete tokens, informed by text embedding from a large pre-trained language model. This method allows for fine-grained language understanding and diverse image editing applications without additional fine-tuning, such as inpainting and mask-free editing.
In the realm of cost-effective and environmentally conscious training, Chen et al.~\cite{chen2023pixart} present PIXART-$\alpha$, a Transformer-based diffusion model that significantly reduces training time and costs while maintaining competitive image quality. Through a decomposed training strategy, efficient text-to-image Transformer design, and higher informative data, PIXART-$\alpha$ demonstrates superior speed, saving resources and minimizing CO2 emissions. It provides a template for startups and the AI community to build high-quality, low-cost generative models.
Lastly, He et al.~\cite{he2023scalecrafter} delve into higher-resolution visual generation with ScaleCrafter, an approach that addresses the challenges of object repetition and structure in images created at resolutions beyond those of the training datasets. By re-dilating convolutional perception fields and implementing dispersed convolution and noise-damped classifier-free guidance, ScaleCrafter enables the generation of ultra-high-resolution images without additional training or optimization, setting a new standard for texture detail and resolution in synthesized images.
Collectively, these advancements represent a paradigm shift in the quality enhancement of diffusion-based generative models, offering innovative solutions to meet the ever-growing demands for high-quality, customizable, and efficient image generation and editing in the AI-powered creative landscape.

\textbf{Specialized Tasks.}
The design and application of diffusion-based methods have extended into specialized tasks, revealing both the potential and the challenges associated with these powerful generative tools.
Somepalli et al.~\cite{somepalli2023diffusion} raise concerns about the originality of the content produced by diffusion models, particularly questioning whether these models generate unique art or merely replicate existing training data. Through image retrieval frameworks, they analyze content replication rates in models like Stable Diffusion and stress the significance of diverse and extensive training sets to mitigate direct copying.
Zhang et al.~\cite{zhang2023prospect} tackle the limitation of personalizing specific visual attributes in generative models. Introducing ProSpect, they utilize the stepwise generation process of diffusion models to represent images with inverted textual token embeddings, corresponding to different stages of image synthesis. This method enhances disentanglement and controllability, enabling attribute-aware personalization in image generation without the need for fine-tuning the diffusion models.
In the realm of vector graphics, Jain et al.~\cite{jain2023vectorfusion} demonstrate that text-conditioned diffusion models trained on pixel representations can be adapted to produce SVG-format vector graphics. Through Score Distillation Sampling loss and a differentiable vector graphics rasterizer, VectorFusion abstracts semantic knowledge from pretrained diffusion models, yielding coherent vector graphics suitable for scalable design applications.
Zhang et al.~\cite{zhang2024transparent} introduce LayerDiffusion, an innovative approach that equips large-scale pretrained latent diffusion models with the capability to generate transparent images and image layers. By incorporating ``latent transparency'' into the model's latent space, LayerDiffusion maintains the quality of the original diffusion model while enabling transparency, facilitating applications like layer generation and structural content control.
These specialized applications of diffusion-based methods highlight the versatility of generative AI, addressing the need for authenticity in digital art, personalization of visual attributes, scalability in design formats, and transparency in image layers. As these technologies advance, they promise to reshape the landscape of digital content creation, offering tools that can adapt to an array of specialized tasks while preserving the integrity and quality of the generated materials.

\begin{table}[h]
\centering
\caption{Method Categorization by Task.}
\begin{tabular}{p{3.2cm}p{11.3cm}}
\toprule
Task & Method \\
\midrule
Controllable Generation & ILVR \cite{choi2021ilvr}, Textual Inversion \cite{gal2022image}, ControlNet \cite{zhang2023adding}, Uni-ControlNet \cite{zhao2024uni}, DreamBooth \cite{ruiz2023dreambooth}, RPG framework \cite{yang2024mastering}, PHDiffusion \cite{lu2023painterly} \\
Content Editing & GLIDE \cite{nichol2021glide}, Prompt-to-Prompt \cite{hertz2022prompt}, Custom Diffusion \cite{kumari2023multi}, InstructPix2Pix \cite{brooks2023instructpix2pix}, pix2pix-zero \cite{parmar2023zero} \\
Stylization & InST \cite{zhang2023inversion}, DiffStyler \cite{huang2024diffstyler}, DreamStyler \cite{ahn2024dreamstyler}, StyleDrop \cite{sohn2024styledrop} \\
Quality Enhancement & eDiff-I \cite{balaji2022ediff}, Muse \cite{chang2023muse}, PIXART-$\alpha$ \cite{chen2023pixart}, ScaleCrafter \cite{he2023scalecrafter} \\
Specialized Tasks & ProSpect \cite{zhang2023prospect}, VectorFusion \cite{jain2023vectorfusion}, LayerDiffusion \cite{zhang2024transparent}, Diffusion Model Originality~\cite{somepalli2023diffusion} \\
\bottomrule
\end{tabular}
\label{tab: task_method}
\end{table}

In Table~\ref{tab: task_method} we summarize the discussed research and provide more examples, to establish correspondence between different generative tasks and methods. 

\subsubsection{Method Classification by Diffusion Model Structure.}

Based on Sec.~\ref{sec: model_structure} and Fig.~\ref{fig:model_structure}, we classify different methods to design or refine diffusion-based models by a unified model structure and summarize representative methods to optimize each module.

\textbf{Encoder-decoder:} 
Lu et al.~\cite{lu2023painterly} innovate with a dual encoder setup in their PHDiffusion model for painterly image harmonization, which features a lightweight adaptive encoder and a Dual Encoder Fusion (DEF) module, allowing for a more nuanced manipulation of foreground features to blend photographic objects into paintings seamlessly.
Yang et al.~\cite{yang2024mastering} push the boundaries of text-to-image diffusion models by introducing the RPG framework, which leverages the complex reasoning capabilities of multimodal LLMs. This model employs a global planner that decomposes the image generation task into sub-tasks, enhancing the model's ability to handle prompts with multiple objects and intricate relationships.

\textbf{Denoiser:}
Liu et al.~\cite{liu2022compositional} propose a compositional visual generation technique that interprets diffusion models as energy-based models. This allows for the combination of multiple diffusion processes, each representing different components of an image, enabling the generation of scenes with a level of complexity not encountered during training.
Bar et al.~\cite{bar2023multidiffusion} present MultiDiffusion, a framework that fuses multiple diffusion paths for controlled image generation. The key innovation lies in its optimization task that allows for high-quality, diverse image output without requiring re-training or fine-tuning.

\textbf{Noise Predictor:}
Chefer et al.~\cite{chefer2023attend} introduce an attention-based semantic guidance system, Attend-and-Excite, for text-to-image diffusion models. This method refines the cross-attention units during inference time, ensuring that generated images more faithfully represent the text prompt's content.
Cao et al.~\cite{cao2023masactrl} develop a tuning-free image synthesis and editing approach, MasaCtrl, which transforms self-attention in diffusion models into mutual self-attention. This allows for consistent generation and editing by querying correlated local content and texture from source images.

\textbf{Additional modules:}
Hu et al.~\cite{hu2021lora} innovate in the adaptation of large language models through LoRA, which introduces low-rank matrices into the Transformer architecture, significantly reducing the number of trainable parameters required for downstream tasks.
Mou et al.~\cite{mou2024t2i} create T2I-Adapters, specialized modules that enhance the controllability of text-to-image models. These adapters tap into the models' implicit knowledge for more nuanced control over the generation outputs, emphasizing color and structure without retraining the entire model.

\begin{table}[h]
\centering
\caption{Method Categorization by Module}
\begin{tabular}{p{4cm}p{10cm}}
\toprule
Module & Method \\
\midrule
Encoder-Decoder & RPG framework \cite{yang2024mastering}, PHDiffusion \cite{lu2023painterly} \\
Denoiser & ILVR \cite{choi2021ilvr}, Compositional Generation\cite{liu2022compositional}, MultiDiffusion \cite{bar2023multidiffusion}, GLIDE \cite{nichol2021glide}, Custom Diffusion \cite{kumari2023multi}, InstructPix2Pix \cite{brooks2023instructpix2pix}, pix2pix-zero \cite{parmar2023zero} \\
Noise Predictor & Attend-and-Excite \cite{chefer2023attend}, MasaCtrl \cite{cao2023masactrl} \\
Additional Modules & LoRA \cite{hu2021lora}, T2I-Adapters \cite{mou2024t2i}, Textual Inversion \cite{gal2022image}, ControlNet \cite{zhang2023adding}, Uni-ControlNet \cite{zhao2024uni}, DreamBooth \cite{ruiz2023dreambooth}, eDiff-I \cite{balaji2022ediff}, Muse \cite{chang2023muse}, PIXART-$\alpha$ \cite{chen2023pixart}, ScaleCrafter \cite{he2023scalecrafter}, ProSpect \cite{zhang2023prospect}, VectorFusion \cite{jain2023vectorfusion}, LayerDiffusion \cite{zhang2024transparent} \\
\bottomrule
\end{tabular}
\label{tab: module_method}
\end{table}

Table~\ref{tab: module_method} illustrates Method Categorization by Model Structure. Each of these innovations contributes significantly to the design and application of diffusion models, enhancing their capacity for a range of generative tasks with improved efficiency, control, and output quality.

\subsubsection{Summary and Trend Identification.}

Following the previous illustration of visual art generative tasks, methods, and diffusion-based model structures, we form them into Table~\ref{tab: module_task} and discuss how they manifest features and trends in diffusion-based method design.

\begin{table}[h]
\centering
\caption{Application of Different Modules in Generative Tasks}
\begin{tabular}{p{2cm}p{3cm}p{3cm}p{2.5cm}p{3cm}}
\toprule
Module & Conditional Generation & Content Editing & Stylization & Quality Enhancement \\
\midrule
Encoder & RPG framework \cite{yang2024mastering}, \newline DreamBooth \cite{ruiz2023dreambooth} & Glide \cite{nichol2021glide}, \newline InstructPix2Pix \cite{brooks2023instructpix2pix} & InST \cite{zhang2023inversion} & eDiff-I \cite{balaji2022ediff} \\
\midrule
Decoder & Textual Inversion \cite{gal2022image}, \newline ControlNet \cite{zhang2023adding} & Glide \cite{nichol2021glide}, \newline InstructPix2Pix \cite{brooks2023instructpix2pix} & DiffStyler \cite{huang2024diffstyler} & PIXART-$\alpha$ \cite{chen2023pixart} \\
\midrule
Denoiser & ILVR \cite{choi2021ilvr}, \newline Uni-ControlNet \cite{zhao2024uni} & Prompt-to-Prompt \cite{hertz2022prompt}, \newline Custom Diffusion \cite{kumari2023multi} & StyleDrop \cite{sohn2024styledrop}, \newline DiffStyler \cite{huang2024diffstyler} & ScaleCrafter \cite{he2023scalecrafter} \\
\midrule
Noise Predictor & - & - & DreamStyler \cite{ahn2024dreamstyler} & Muse \cite{chang2023muse} \\
\midrule
Additional Modules & LoRA \cite{hu2021lora}, \newline Textual Inversion \cite{gal2022image} & pix2pix-zero \cite{parmar2023zero}, \newline T2I-Adapters \cite{mou2024t2i} & StyleDrop \cite{sohn2024styledrop} & VectorFusion \cite{jain2023vectorfusion}, \newline LayerDiffusion \cite{zhang2024transparent} \\
\bottomrule
\end{tabular}
\label{tab: module_task}
\end{table}

This table has multiple implications. In designing a method for specific generative tasks, we may start from a column and select different corresponding modules to test their performance. We may also combine modules from different columns, which may help us accomplish multiple tasks simultaneously. On the other hand, we summarize the following trends in adapting diffusion modules and designing methods for visual art creation:
\begin{itemize}
  \item \textbf{Integration of Attention Mechanisms:} The trend towards incorporating sophisticated attention mechanisms~\cite{vaswani2017attention, zhang2019self} within generative models is evident, allowing for more detailed and contextually relevant image generation \cite{chefer2023attend, cao2023masactrl}.
  \item \textbf{Enhanced Personalization and Fine-tuning:} Techniques for fine-tuning pre-trained models to adapt to specific styles, subjects, or user preferences with minimal computation are gaining traction \cite{gal2022image, ruiz2023dreambooth}.
  \item \textbf{Control and Precision in Content Generation:} Methods have been developed for precise control over layout, style, and content, indicating an increased focus on user-guided generation \cite{choi2021ilvr, zhang2023adding}.
  \item \textbf{Quality Enhancement through Advanced Training and Loss Functions:} Innovations in training strategies and loss function designs aim to produce high-fidelity outputs \cite{balaji2022ediff, he2023scalecrafter, wang2022learning}.
  \item \textbf{Modularity and Composability in Model Design:} The design of modular and composable components reflects a trend toward more adaptable generative systems \cite{zhao2024uni, hu2021lora, mou2024t2i}.
  \item \textbf{Multi-Task and Multi-Modal Generative Models:} The development of models capable of handling multiple tasks or modalities points to a trend towards versatile models \cite{kumari2023multi, brooks2023instructpix2pix, zhang2023prospect}.
  \item \textbf{Efficiency and Scalability:} Innovations in model architecture aim to enhance the generation process while ensuring computational efficiency \cite{chen2023pixart, jain2023vectorfusion}.
\end{itemize}

\section{Discussion}
\label{sec: discussion}
In this section, we focus on the frontiers, trends, and future work of Diffusion-Based Visual Art Creation (\textbf{Q4}). Specifically, we adopt a technical and synergistic perspective to better characterize the multidimensional essence of this interdisciplinary field. In this way, we aim to shed light on emerging topics and possible future developments to provide inspiration and guidance for scientific researchers, artistic practitioners, and the whole community (\textbf{G2}).

\subsection{Breaking the Fourth Wall: a Technical Perspective}

The first trend is facilitating the creation of a more artistic, controllable, and realistic environment through the transcendence of dimensions. Researchers combine higher-dimension visual content and more diverse modalities with advanced computational power to create an immersive experience.

\subsubsection{Higher Dimension.}
% 3D  cases (figure)——
% Fast generation method of 3D scene in Chinese landscape painting

The first series of work revolve around the innovative integration of AI with 3D artistic expression and scene generation. Among them, ARF~\cite{zhang2022arf} presents a method to transfer artistic features from a 2D style image to a 3D scene, using a radiance field representation that overcomes geometric reconstruction errors found in previous techniques. It introduces a nearest neighbor-based loss for capturing style details and a deferred back-propagation method, optimizing memory-intensive radiance fields and improving the visual quality of stylized scenes.
CoARF~\cite{zhang2024coarf} builds on this by introducing a novel algorithm for controllable 3D scene stylization. It offers fine-grained control over the style transfer process using segmentation masks with label-dependent loss functions, and a semantic-aware nearest neighbor matching algorithm, achieving superior style transfer quality.
Instruct-NeRF2NeRF~\cite{haque2023instruct} proposes a method for editing NeRF scenes with text instructions, using an image-conditioned diffusion model to achieve realistic targeted edits. The technique allows large-scale, real-world scene edits, expanding the possibilities for user-driven 3D content creation.
DreamWire~\cite{qu2023wired} presents an AI system for crafting multi-view wire art, using a combination of 3D Bézier curves, Prim's algorithm, and knowledge distillation from diffusion models. This system democratizes the creation of multi-view wire art (MVWA), making it accessible to non-experts while ensuring visual aesthetics.
Lastly, RealmDreamer~\cite{shriram2024realmdreamer} introduces a technique for text-driven 3D scene generation using 3D Gaussian Splatting and image-conditional diffusion models. It uniquely generates high-quality 3D scenes in diverse styles without the need for video or multi-view data, showcasing the potential for 3D synthesis from single images.
Together, these papers advance the fusion of generative AI and 3D art, enabling new levels of creativity and control in digital scene creation and artistic expression.

\subsubsection{Diverse Modalities.}
% dynamic and motion cases——integration of more modalities
The second series of works showcase innovations that span across visual and auditory domains, aligning technologies with the nuanced dynamics of human perception and artistic creation.
The Human-Art dataset~\cite{ju2023human} addresses a void in computer vision by collating 50k images from both natural and artificial human depictions across 20 different scenarios, marking a leap forward in human pose estimation and image generation tasks. This versatile dataset, with over 123k annotated person instances in both 2D and 3D, stands to offer new insights and research directions as it bridges the gap between natural and artificial scenes.
SonicDiffusion~\cite{biner2024sonicdiffusion} introduces an audio-driven approach to image generation and editing, leveraging the multimodal aspect of human perception. By translating audio features into tokens compatible with diffusion models, and incorporating audio-image cross-attention layers, SonicDiffusion demonstrates superior performance in creating and editing images conditioned on auditory inputs.
Sprite-from-Sprite~\cite{zhang2022sprite} unravels the complexity of cartoon animations by decomposing them into basic "sprites" using a pioneering self-supervised framework that leverages Pixel MLPs. This method cleverly simplifies the decomposition of intricate animated content by first resolving simpler sprites, thus easing the overall process and enhancing the quality of cartoon animation analysis.
WonderJourney~\cite{yu2023wonderjourney} transforms scene generation by introducing a modularized framework designed to create a perpetual sequence of diverse and interconnected 3D scenes from any starting point, be it a textual description or an image. This approach yields imaginative and visually diverse scene sequences, showcasing the framework's robust versatility.
Lastly, Intelligent-paint~\cite{wang2024intelligent} propels the generation of Chinese painting processes forward. Utilizing a Vision Transformer (ViT)-based generator, adversarial learning, and loss constraints that adhere to the characteristics of Chinese painting, this method vastly improves the plausibility and clarity of intermediate painting steps. The approach not only successfully bridges the gap between generated sequences and real painting processes but also serves as a valuable learning tool for novices in the art of Chinese painting.
Collectively, these contributions present a multifaceted view of the convergence between AI technologies and the arts, pushing the boundaries of what can be achieved in terms of human-centric data analysis, multimodal synthesis, and artistic process generation.

\subsection{``1 + 1 > 2'': a Synergistic Perspective}

The second trend is to promote human and AI's understanding and collaboration with each other, and finally to unleash human potential and stimulate creativity in diffusion-based visual art creation. Research under this topic are mostly understanding-oriented and application-driven, including creative system design, multiple intuitive interactions, content reception and modality alignment. We summarize different approaches and solutions for the problems and tasks.

\subsubsection{Interactive Systems.}
% HCI cases (human computer interaction) and metaverse (figure)

Emerging research showcases interactive technologies that amalgamate human intuition with AI's capabilities to enhance the process of creation across various artistic domains.
PromptPaint~\cite{chung2023promptpaint} revolutionizes text-to-image models by allowing users to intuitively guide image generation through paint-medium-like interactions, akin to mixing colors on a palette. This system enables the iterative application of prompts to canvas areas, enhancing the user's ability to shape outputs in ways that language alone could not facilitate.
Collaborative Neural Painting~\cite{dall2023collaborative} introduces the task of joint art creation between humans and AI. Through a novel Transformer-based architecture that models the input and completion strokes, users can iteratively shape the artwork, making the painting process both creative and collaborative.
ArtVerse~\cite{guo2023artverse} proposes a human-machine collaborative creation paradigm in the metaverse, where AI participates in artistic exploration and evolution, shaping a decentralized ecosystem for art creation, dissemination, and transaction.
ARtVista~\cite{hoang2024artvista} empowers individuals to bridge the gap between conceptual ideas and their visual representation. By integrating AR and generative AI, ARtVista assists users in creating sketches from abstract thoughts and generating vibrant paintings in diverse styles. Its unique paint-by-number simulation further simplifies the artistic process, enabling anyone to produce stunning artwork without advanced drawing skills.
Interactive3D~\cite{dong2024interactive3d} framework elevates 3D object generation by granting users unparalleled control over the creation process. Utilizing Gaussian Splatting and InstantNGP representations, this framework allows for comprehensive interaction, including adding, removing, transforming, and detailed refinement of 3D components, pushing the boundaries of precision in generative 3D modeling.
Finally, Neural Canvas~\cite{shen2024NeuralCanvas} integrates generative AI into a 3D sketching interface, facilitating scenic design prototyping. It transcends the limitations of traditional tools by enabling rapid iteration of visual ideas and atmospheres in 3D space, expediting the design process for both novices and professionals.
These contributions collectively demonstrate a synergistic approach where the sum of collaborative human and machine efforts yields greater creative outcomes than either could achieve independently, marking a new era in interactive and generative art-making.

\subsubsection{Reception and Alignment.}
Another series of work focus on latent space disentanglement and multi-modality alignment, combining the perspectives of content reception and generation for understanding, to enable human and AI better understand each other.
For example, a study on multi-sensory experience~\cite{cho2021study} emphasizes the potential of various sensory elements such as sound, touch, and smell to convey visual artwork elements to visually impaired individuals. By leveraging patterns, temperature, and other sensory cues, this research opens up new avenues for inclusive art appreciation and paves the way for further exploration in multi-sensory interfaces.
In the realm of knowledgeable art description, a new framework~\cite{bai2021explain} has been introduced for generating rich descriptions of paintings that cover artistic styles, content, and historical context. This multi-topic approach, augmented with external knowledge, has been shown to successfully capture diverse aspects of artwork and its creation, enhancing the viewer's understanding and engagement with art.
Initial Images~\cite{qiao2022initial} explores the use of image prompts alongside text to improve the subject representation in AI-generated art. This research demonstrates how image prompts can exert significant control over final compositions, leading to more accurate and user-aligned creations.
CLIP-PAE~\cite{zhou2023clip} addresses the challenge of disentangled and interpretable text-guided image manipulation by introducing projection-augmentation embedding. This method refines the alignment between text and image features, enabling more precise and controllable manipulations, particularly demonstrated in the context of facial editing.
Evaluating text-to-visual generation has been advanced with the introduction of VQAScore~\cite{lin2024evaluating}, a metric that utilizes a visual-question-answering model to assess image-text alignment. This approach offers a more nuanced evaluation of complex prompts and has led to the creation of GenAI-Bench, a benchmark for rigorously testing generative AI models against compositional text prompts.
Collectively, these contributions signify a synergistic advancement where the combination of multiple approaches, senses, and technologies results in a more profound and aligned interaction between AI and human perception, pushing the boundaries of art creation, appreciation, and evaluation.

% figure
\begin{figure}
    \centering
    \includegraphics[width=0.6\textwidth]{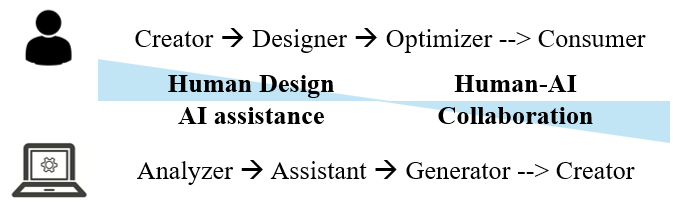}
    \caption{New perspectives on the emerging frontiers and future work of diffusion-based visual art creation. From a synergistic perspective, we inspect a continuous paradigm shift in the roles of human and AI.
    }
    \label{fig:paradigm} 
\end{figure}

Fig.~\ref{fig:paradigm} illustrates the paradigm shift in Human and AI’s roles in content creation. With technological advancements, human roles in AIGC are shifting from creators to optimizers to consumers, while AI develops from analyzers to generators to creators. In all, the creative paradigm shifts from human design and AI assistance to human-AI collaboration, where the two counterparts learn from and inspire each other.

\section{Conclusion}

This survey has charted the course of diffusion-based generative methods within the rich terrain of visual art creation. We began by identifying the research scope, pinpointing diffusion models and visual art as pivotal concepts, and outlining our dual research goals and the quartet of research questions. A robust dataset was assembled, encompassing relevant papers that underwent a rigorous four-phase filtering process, leading to their categorization across seven dimensions within the thematic angles of application, understanding, and generation.
Through a blend of structural and temporal analysis, we discovered prevailing trends and constructed a comprehensive analytical framework. Our synthesis of findings crystallized into a paradigm, encompassing the quadrants of scenario, modality, task, and method, which collectively shape the nexus of diffusion-based visual art creation.
% Diving deeper into the confluence of technology and artistry, we spotlighted two burgeoning topics. Firstly, the integration and manipulation of human concepts to not only enhance generative outcomes but also to achieve specific visual effects. Secondly, the emergence of human-and-AI inspired model design, which is steered by psychological insights into human perception, mapping the trajectory towards more empathetic and cognitively aligned creations.
As we gaze into the future, we propose a novel perspective that intertwines technological and synergistic aspects, characterizing the collaborative ventures between humans and AI in creating visual art. This perspective beckons a future where AI not only complements human artistry but also actively contributes to the creative process. 
However, amidst the rapid technical strides, we are compelled to ponder the implications of AI potentially surpassing human capacity in both understanding and task execution. In such a scenario, we are prompted to question the pursuits we should embrace. If human aspirations and desires continue to expand, how can AI evolve to meet these ever-growing needs? How can we ensure that the evolution of AI in visual art creation remains aligned with human values and creative aspirations?
In conclusion, while we acknowledge the remarkable progress made thus far, this survey also serves as a clarion call to the research community. As the horizon of AI in visual art creation broadens, we must continue to explore, innovate, and critically reflect on the role of AI in this field. The future beckons with a promise of AI that not only mimics but enriches human creativity, forming an indelible part of our artistic and cultural expression.

%  In this part, we aim to answer: 1)
% What are the specific requirements of AI generated visual art? 2) How are they reflected in the generative goal and evaluation metrics?
% After identifying the interdisciplinary issues of Diffusion-Based 

% \begin{acks}
% \end{acks}

\bibliographystyle{ACM-Reference-Format}
\bibliography{main}


\end{document}